# Dual Modelling of Permutation and Injection Problems

**Brahim Hnich** BRAHIM@4C.UCC.IE
*Cork Constraint Computation Center*
*University College Cork, Cork, Ireland*

**Barbara M. Smith** B.M.SMITH@HUD.AC.UK
*School of Computing and Engineering*
*Huddersfield, U.K*

**Toby Walsh** TW@4C.UCC.IE
*Cork Constraint Computation Center*
*University College Cork, Cork, Ireland.*

## Abstract

When writing a constraint program, we have to choose which variables should be the decision variables, and how to represent the constraints on these variables. In many cases, there is considerable choice for the decision variables. Consider, for example, permutation problems in which we have as many values as variables, and each variable takes an unique value. In such problems, we can choose between a primal and a dual viewpoint. In the dual viewpoint, each dual variable represents one of the primal values, whilst each dual value represents one of the primal variables. Alternatively, by means of channelling constraints to link the primal and dual variables, we can have a combined model with both sets of variables. In this paper, we perform an extensive theoretical and empirical study of such primal, dual and combined models for two classes of problems: permutation problems and injection problems. Our results show that it often be advantageous to use multiple viewpoints, and to have constraints which channel between them to maintain consistency. They also illustrate a general methodology for comparing different constraint models.

## 1. Introduction

Constraint programming is a highly successful technology for solving a wide variety of combinatorial problems like resource allocation, transportation, and scheduling. A constraint program consists of a set of decision variables, each with an associated domain of values, and a set of constraints defining allowed values for subsets of these variables. The efficiency of a constraint program depends on many factors including a good choice for the decision variables, and careful modelling of the constraints on these variables. There is often considerable choice as to what the decision variables and their values should represent. For example, in an exam timetabling problem, the variables could represent the exams, and the values represent the times. Alternatively, we can use a *dual* model in which the variables are the times, and the values are the exams. We always have a choice of this kind in permutation problems. In a permutation problem, we have as many values as variables, and each variable takes an unique value. We can therefore easily exchange the roles of the variables and the values in representing the underlying problem. Many assignment, scheduling and routing problems are permutation problems. For example, sports tournament scheduling can be modelled as finding a permutation of the games to fit into the time slots, or a permutation of the time slots to fit the games into. The aim of this paper is to compare such different models both theoretically and empirically.

The paper is structured as follows. In Section 2, we give the formalism and notation used in the rest of the paper. In Section 3, we present Langford's problem, which is used to illustrate the different ways we can model a permutation problem. We then introduce a formal measure of constraint tightness (Section 4) used to compare theoretically the different models of permutation problems (Section 5). In Section 6, we compare SAT (Boolean) models of permutation problems. In Sections 7 and 8, we complement the theoretical results with some asymptotic and experimental analysis. We then explore the benefits to branching heuristics of having multiple viewpoints of the permutation (section 9). In Section 10, we extend our analysis to injective mappings. Finally, we end with related work (Section 11) and conclusions (Section 12).

## 2. Formal Background

A *constraint satisfaction problem* (CSP) is a set of variables, each with a finite domain of values, and a set of constraints. A constraint consists of a list of variables (the *scope*) and a relation defining the allowed values for these variables. A binary constraint is a constraint whose scope is a pair of variables. A solution to a constraint satisfaction problem is an assignment of values to variables that satisfies all the constraints.

A *permutation problem* is a constraint satisfaction problem in which each decision variable takes an unique value, and there is the same number of values as variables. Hence any solution assigns a permutation of the values to the variables. An important feature of permutation problems is that we can transpose the roles of the variables and the values in representing the underlying problem to give a new *dual* model which is also a permutation problem. Each variable in the original (*primal*) CSP becomes a value in the dual CSP, and vice versa. The primal and the dual CSPs are equivalent since any solution to one can be translated into a solution to the other.

We can choose either model arbitrarily to be the primal model, although in practice it might be easier to express the problem constraints in one of the models rather than the other, so we might tend to think of that model as the primal. We also consider *multiple permutation problems* in which the variables divide into a number of (possibly overlapping) sets, each of which is a permutation problem. This lets us discuss problems like quasigroups. An order $n$ quasigroup (or Latin square) can be modeled as a multiple permutation problem containing $2n$ overlapping permutation problems.

An *injection problem* is a constraint satisfaction problem in which each decision variable takes an unique value, but there are now more values than variables. (Obviously, if there are fewer values than variables, the problem is trivially unsatisfiable.)

Many levels of local consistency have been defined for constraint satisfaction problems involving binary constraints (for references see Debruyne and Bessière, 1997). A problem is $(i, j)$*-consistent* iff it has non-empty domains and any consistent instantiation of $i$ variables can be consistently extended to $j$ additional variables. A problem is *arc-consistent* (AC) iff it is $(1, 1)$-consistent. A problem is *path-consistent* (PC) iff it is $(2, 1)$-consistent. A problem is *strong path-consistent* (ACPC) iff it is AC and PC. A problem is *path inverse consistent* (PIC) iff it is $(1, 2)$-consistent. A problem is *restricted path-consistent* (RPC) iff it is AC and if a value assigned to a variable is consistent with just one value for an adjoining variable then for any other variable there is a compatible value. A problem is *singleton arc-consistent* (SAC) iff it has non-empty domains and for any instantiation of a variable, the resulting subproblem can be made AC.

For non-binary constraints, there has been less work on different levels of local consistency. One exception is generalized arc-consistency. A CSP with binary or non-binary constraints is *generalized arc-consistent* (GAC) iff for any value for a variable in a constraint, there exist compatible values for all the other variables in the constraint. For ordered domains (such as integers), a problem is *bounds consistent* (BC) iff it has non-empty domains and an assignment of its minimum or maximum value to any variable in a (binary or non-binary) constraint can be consistently extended to the other variables in the constraint. In line with the definitions introduced by Debruyne and Bessière (1997), we say that a local consistency property $A$ is as strong as a local consistency property $B$ (written $A \hookrightarrow B$) iff in any problem in which $A$ holds then $B$ holds, $A$ is stronger than $B$ (written $A \rightarrow B$) iff $A \hookrightarrow B$ but not $B \hookrightarrow A$, $A$ is incomparable with $B$ (written $A \otimes B$) iff neither $A \hookrightarrow B$ nor $B \hookrightarrow A$, and $A$ is equivalent to $B$ (written $A \leftrightarrow B$) iff both $A \hookrightarrow B$ and $B \hookrightarrow A$. It has been shown that: ACPC $\rightarrow$ SAC $\rightarrow$ PIC $\rightarrow$ RPC $\rightarrow$ AC $\rightarrow$ BC (Debruyne & Bessière, 1997).

Backtracking algorithms are often used to find solutions to CSPs. Such algorithms try to extend partial assignments, enforcing a local consistency after each extension and backtracking when this local consistency no longer holds. For example, the *forward checking* algorithm (FC) maintains a restricted form of AC that ensures that the binary constraints between the most recently instantiated variable and any uninstantiated variables are AC. FC has been generalized to non-binary constraints (Bessière, Meseguer, Freuder, & Larrosa, 1999). nFC0 makes every $k$-ary constraint with $k-1$ variables instantiated AC. nFC1 applies (one pass of) AC to each constraint or constraint projection involving the current and exactly one future variable. nFC2 applies (one pass of) GAC to each constraint involving the current and at least one future variable. Three other generalizations of FC to non-binary constraints, nFC3 to nFC5, degenerate to nFC2 on the single non-binary constraint describing a permutation, so are not considered here. Finally, the *maintaining arc-consistency* algorithm (MAC) maintains AC during search, whilst MGAC maintains GAC.

## 3. An Example

The $n$-queens problem is one of the simplest examples of a permutation problem. A common and natural model has a decision variable for each row, with its value being the column in which the queen on that row lies. The dual model has a decision variable for each column, with its value being the row on which the queen in that column lies. However, the $n$-queens problem is not combinatorially challenging as it becomes easier as $n$ grows. For example, Morris (1992) has argued that there are no local maxima so throwing queens at random onto the board and performing min-conflicts hill-climbing will almost surely find a solution. We focus therefore on a different permutation problem that is simple like the $n$-queens problem but appears to be more combinatorially challenging. By using a simple example, the characteristics of permutation problems are hopefully more apparent than in more complex problems where the other constraints have a larger impact.

Langford's problem is **prob024** in CSPLib (Gent & Walsh, 1999). A comprehensive history of the problem is given by Miller (2002). The problem is defined as follows:

> *"A 27-digit sequence includes the digits 1 to 9 three times each. There is one digit between the first two 1s, and one digit between the last two 1s. There are just two digits between the first two 2s, and two digits between the last two 2s, ...and so on. Find all possible such sequences."*

The problem can easily be generalized to the $(n, m)$ problem where we have a sequence of length $n * m$, containing the integers 1 to $m$ repeated exactly $n$ times. The above problem is thus the (3,9) problem. It has exactly 6 solutions:

```
181915267285296475384639743
191218246279458634753968357
191618257269258476354938743
347839453674852962752816191
347936483574692582762519181
753869357436854972642812191
```

Note that the last three solutions are the reverse of the first three. This symmetry can be eliminated by adding constraints; for instance, in the (3,9) problem the second 9 cannot be placed in the second half of the sequence, and if it is in the central position in the sequence, the second 8 must be placed in the first half of the sequence. Such constraints have been added in what follows.

The first model of Langford's problem we will consider, which we shall arbitrarily call the primal model, has a variable for each occurrence of the digits. The value of this variable is the position in the sequence of this occurrence. For example, the (3,9) problem has 27 variables, $x_i$ with $i \in [1, 27]$. The value of $x_i$ is the location in the sequence of the $i$ div $m$+1th occurrence of the digit $i$ mod $m$. Thus, $x_1$ has as its value the location of the 1st occurrence of the digit 1, $x_2$ has as its value the location of the 1st occurrence of the digit 2, ..., $x_9$ has as its value the location of the 1st occurrence of the digit 9, $x_{10}$ has as its value the location of the 2nd occurrence of the digit 1, and so on. We have a permutation constraint that ensures that each digit occurrence occurs at a different position in the sequence. This can be implemented either as a global all-different constraint on all the $x_i$, or as pairwise not-equals constraints on each possible pair of variables. We call the former the "primal all-different" model and the later the "primal not-equals" model. Finally, we have constraints that the digit occurrences occur in order down the sequence and constraints on the separation of the different occurrences of a digit: that is we have $x_i < x_{i+m} < x_{i+2m}$, $x_{i+m} - x_i = i$ and $x_{i+2m} - x_{i+m} = i$ for $i \leq m$.

Table 1 gives the primal representation of the sequence `23421314`, a solution to the (2,4) problem. For clarity, we also indicate the corresponding digit occurrence using the notation "$d_k$" for the $k$th occurrence of the digit $d$. For example, $3_2$ is the 2nd occurrence of the digit "3" and $2_1$ is the 1st occurrence of the digit "2".

| Index ($i$) | 1 | 2 | 3 | 4 | 5 | 6 | 7 | 8 |
|---|---|---|---|---|---|---|---|---|
| Value of primal variable ($x_i$) | 5 | 1 | 2 | 3 | 7 | 4 | 6 | 8 |
| Equivalent digit occurrence | $1_1$ | $2_1$ | $3_1$ | $4_1$ | $1_2$ | $2_2$ | $3_2$ | $4_2$ |

Table 1: The primal representation of the sequence `23421314`, a solution of the (2,4) problem.

The dual model of Langford's problem has a variable for each location in the sequence. The value of this variable represents the digit occurrence at this location. For example, the (3,9) problem has 27 variables, $d_j$ with $j \in [1, 27]$. The value $i$ of $d_j$ is an integer in the interval $[1, n*m]$, representing the fact that the $i$ div $m$+1th occurrence of the digit $i$ mod $m$ occurs at location $j$. Thus, $d_3 = 2$ represents the fact that the 1st occurrence of the digit 2 occurs at the 3rd location, and $d_4 = 10$ represents the fact that the 2nd occurrence of the digit 1 occurs at the 4th location, and so on.

In the dual model, we again have a permutation constraint that each location contains a different digit occurrence. This can again be implemented via a global all-different constraint on the $d_j$ or by pairwise not-equals constraints on each pair of dual variables. We call the former the "dual all-different" model and the later the "dual not-equals" model. The separation constraints are not as simple to specify in the dual model. For example, for $i \leq m$, we can add constraints of the form: $d_j = i$ iff $d_{j+i+1} = i + m$ and $d_j = i$ iff $d_{j+2*(i+1)} = i + 2 * m$. Table 2 gives the dual representation of the sequence `23421314`, a solution to the (2,4) problem.

| Index ($j$) | 1 | 2 | 3 | 4 | 4 | 6 | 7 | 8 |
|---|---|---|---|---|---|---|---|---|
| Value of dual variable ($d_j$) | 2 | 3 | 4 | 6 | 1 | 7 | 5 | 8 |
| Equivalent digit occurrence | $2_1$ | $3_1$ | $4_1$ | $2_2$ | $1_1$ | $3_2$ | $1_2$ | $4_2$ |

Table 2: Dual representation of the sequence `23421314`, a solution of the (2,4) problem.

It is possible to combine primal and dual models by linking the two sets of variables, using *channelling constraints* to maintain consistency between the two viewpoints. This approach is called "redundant modelling" by Cheng et al. (1999). A similar idea was previously suggested, specifically for permutation problems, by Geelen (1992). In Langford's problem, the channelling constraints are $x_i = j$ iff $d_j = i$, and constraints of the same form can be used in building a combined primal/dual model of any permutation problem. Many constraint toolkits support channelling of this kind with efficient global constraints. For example, ILOG Solver has a constraint, `IlcInverse`, which can be used to replace a set of individual constraints of the form $x_i = j$ iff $d_j = i$, and the Sicstus finite domain constraint library has an `assignment` predicate which can be used similarly.

The combined model is clearly redundant as we can delete the constraints of either individual model without increasing the set of solutions. For instance, in Langford's problem, we need only express the separation constraints in terms of either the primal or the dual variables. More surprisingly, the permutation constraints on both the primal and the dual variables are also redundant. The existence of the dual variables and the channelling constraints linking them to the primal variables are sufficient to ensure that the values assigned to the primal variables are a permutation (and therefore the same must be true of the dual variables).

Even if constraints are logically redundant (that is, they can be deleted without changing the set of solutions), they may still be useful during search. Logically redundant constraints are often called "implied constraints", and useful implied constraints are frequently added to a model to increase the amount of constraint propagation (Smith, Stergiou, & Walsh, 2000)). In the next section, we present a measure of constraint tightness that allows us to determine when an implied constraint added to a model will improve constraint propagation. In the following section, we apply this measure of constraint tightness to the different models of permutation problems introduced in this section. We are able to show, for example, that the channelling constraints not only make the binary not-equals constraints redundant: they are tighter and can give more domain pruning.

## 4. Constraint Tightness

Our definition of constraint tightness assumes that constraints are defined over the same variables and values or, as in the case of primal and dual models, variables and values which are bijectively related. In this way, we can always compare like with like. Our definition of constraint tightness is strongly influenced by the way local consistency properties are compared by Debruyne and Bessière (1997). Indeed, the definition is parameterized by a local consistency property since the amount of pruning provided by a set of constraints depends upon the level of local consistency being enforced. If we enforce a high level of local consistency, we may get as much constraint propagation with a loose constraint as a much lower level of local consistency applied to a tight constraint. Our measure of constraint tightness would also be useful in a number of other applications (e.g. reasoning about the impact of different local consistency techniques on a single fixed model).

Consider a set of constraints $A$ defined over a set of variables $V_A$, and another set of constraints $B$ defined over a set of variables $V_B$, where there is a bijection between assignments to $V_A$ and $V_B$ (in the rest of the paper, this bijection is either the identity map, or that defined by the channelling constraints). We say that the set of constraints $A$ is *at least as tight as* the set $B$ with respect to $\Phi$-consistency (written $\Phi_A \hookrightarrow \Phi_B$) iff, given any domains for their variables, if $A$ is $\Phi$-consistent then the equivalent domains of $B$ according to the bijection are also $\Phi$-consistent. By considering all possible domains for the variables, this ordering measures the potential for domains to be pruned during search as variables are instantiated and domains pruned (possibly by other constraints in the problem). Note that we discuss the equivalent domains so that we can consider primal and dual models in which the variables and values are different (but are in one to one relation with each other). We say that a set of constraints $A$ is *tighter* than a set $B$ wrt $\Phi$-consistency (written $\Phi_A \rightarrow \Phi_B$) iff $\Phi_A \hookrightarrow \Phi_B$ but not $\Phi_B \hookrightarrow \Phi_A$, $A$ is *incomparable* to $B$ wrt $\Phi$-consistency (written $\Phi_A \otimes \Phi_B$) iff neither $\Phi_A \hookrightarrow \Phi_B$ nor $\Phi_B \hookrightarrow \Phi_A$, and $A$ is *equivalent* to $B$ wrt $\Phi$-consistency (written $\Phi_A \leftrightarrow \Phi_B$) iff both $\Phi_A \hookrightarrow \Phi_B$ and $\Phi_B \hookrightarrow \Phi_A$. We can easily generalize these definitions to compare Φ-consistency on $A$ with Θ-consistency on $B$. This definition of constraint tightness has some nice monotonicity and fixed-point properties which we will use extensively throughout this paper.

**Property 1 (monotonicity and fixed-point)**

1. $AC_{A\cup B} \hookrightarrow AC_A \hookrightarrow AC_{A\cap B}$
2. $AC_A \to AC_B$ *implies* $AC_{A\cup B} \leftrightarrow AC_A$

Similar monotonicity and fixed-point properties hold for BC, RPC, PIC, SAC, ACPC, and GAC. We also extend these definitions to compare constraint tightness wrt search algorithms like MAC and FC that maintain some local consistency during search. For example, we say that $A$ is *at least as tight as* $B$ wrt algorithm $X$ (written $X_A \hookrightarrow X_B$) iff, given any fixed variable and value ordering and any domains for the variables of $A$, $X$ visits no more nodes to find a solution of $A$ or prove it unsatisfiable than $X$ visits on $B$ with the equivalent domains, and the equivalent variable and value ordering. Equivalence here is again with respect to the bijection between the assignments to the variables of $A$ and to $B$. We say that $A$ is *tighter* than $B$ wrt algorithm $X$ (written $X_A \to X_B$) iff $X_A \hookrightarrow X_B$ but not $X_B \hookrightarrow X_A$. Similar monotonicity and fixed-point properties can be given for FC, MAC and MGAC. Finally, we write $X_A \Rightarrow X_B$ if $X_A \to X_B$ and there is a parameterized set of problems of size $n$ and a fixed variable and value ordering with which $X$ visits exponentially fewer nodes in $n$ when applied to $A$ than when applied to $B$. Our results can be extended to algorithms that find all solutions. In addition, they can also be extended to a restricted class of dynamic variable and value orderings (Bacchus, Chen, van Beek, & Walsh, 2002).

## 5. Theoretical Comparison

We now have the theoretical machinery needed to compare the different ways we can model a permutation problem such as Langford's problem. The *primal* not-equals model of a permutation has not-equals constraints between the variables in each permutation. The *primal* all-different model has an all-different constraint between the variables in each permutation. In a *dual* model, we interchange variables for values. A combined *primal and dual* model has both the primal and the dual variables, and *channelling constraints* linking them, of the form: $x_i = j$ iff $d_j = i$ where $x_i$ is a primal variable and $d_j$ is a dual variable. A combined model can also have not-equals and/or all-different constraints on the primal and/or dual variables. There will, of course, typically be other constraints on both sets of variables which depend on the nature of the permutation problem. For example, in Langford's problem we also have the separation constraints. As a second example, in the all-interval series problem from CSPLib, the variables and the differences between neighboring variables are both permutations. In what follows, we do not consider directly the contribution of such additional constraints to pruning. However, the ease with which we can express each additional constraint in the primal or the dual model and the resulting pruning power of these constraints may determine our choice of the primal, dual or combined model.

We will use the following subscripts: "$\neq$" for the primal not-equals constraints, "$c$" for channelling constraints, "$\neq c$" for the primal not-equals and channelling constraints, "$\neq c \neq$" for the primal not-equals, dual not-equals and channelling constraints, "$\forall$" for the primal all-different constraint, "$\forall c$" for the primal all-different and channelling constraints, and "$\forall c \forall$" for the primal all-different, dual all-different and channelling constraints. Thus AC$_{\neq}$ is AC applied to the primal not-equals constraints, whilst SAC$_{\neq c}$ is SAC applied to the primal not-equals and channelling constraints.

### 5.1 Arc-Consistency

We first prove that, with respect to AC, channelling constraints are tighter than the primal not-equals constraints, but less tight than the primal all-different constraint.

**Theorem 1** *On a permutation problem:*

$$GAC_{\forall c\forall} \leftrightarrow GAC_{\forall c} \leftrightarrow GAC_{\forall} \rightarrow AC_{\neq c\neq} \leftrightarrow AC_{\neq c} \leftrightarrow AC_{c} \rightarrow AC_{\neq}$$

**Proof:** In this and following proofs, we just prove the most important results. Others follow quickly, often using transitivity, monotonicity and the fixed-point theorems.

To show GAC$_{\forall}$ $\rightarrow$ AC$_c$, consider a permutation problem whose primal all-different constraint is GAC. Suppose the channelling constraint between $x_i$ and $d_j$ was not AC. Then either $x_i$ is set to $j$ and $d_j$ has $i$ eliminated from its domain, or $d_j$ is set to $i$ and $x_i$ has $j$ eliminated from its domain. But neither of these two cases is possible by the construction of the primal and dual model. Hence the channelling constraints are all AC. To show strictness, consider a 5-variable permutation problem in which $x_1 = x_2 = x_3 = \{1, 2\}$ and $x_4 = x_5 = \{3, 4, 5\}$. This is AC$_c$ but not GAC$_{\forall}$.

To show AC$_c$ $\rightarrow$ AC$_{\neq}$, suppose that the channelling constraints are AC. Consider a not-equals constraint, $x_i \neq x_j$ $(i \neq j)$ that is not AC. Now, $x_i$ and $x_j$ must have the same singleton domain, $\{k\}$. Consider the channelling constraint between $x_i$ and $d_k$. The only AC value for $d_k$ is $i$. Similarly, the only AC value for $d_k$ in the channelling constraint between $x_j$ and $d_k$ is $j$. But $i \neq j$. Hence, $d_k$ has no AC values. This is a contradiction as the channelling constraints are AC. Hence all not-equals constraints are AC. To show strictness, consider a 3-variable permutation problem with $x_1 = x_2 = \{1, 2\}$ and $x_3 = \{1, 2, 3\}$. This is AC$_{\neq}$ but is not AC$_c$.

To show AC$_{\neq c\neq}$ $\leftrightarrow$ AC$_c$, by monotonicity, AC$_{\neq c\neq}$ $\hookrightarrow$ AC$_c$. To show the reverse, consider a permutation problem which is AC$_c$ but not AC$_{\neq c\neq}$. Then there exists at least one not-equals constraint that is not AC. Without loss of generality, let this be on two dual variables (a symmetric argument can be made for two primal variables). So both the associated (dual) variables, call them $d_i$ and $d_j$ must have the same singleton domain, say $\{k\}$. Hence, the domain of the primal variable $x_k$ includes $i$ and $j$. Consider the channelling constraint between $x_k$ and $d_i$. Now this is not AC as the value $x_k = j$ has no support. This is a contradiction.

To show GAC$_{\forall c\forall}$ $\leftrightarrow$ GAC$_{\forall}$, consider a permutation problem that is GAC$_{\forall}$. For every possible assignment of a value to a variable, there exist a consistent extension to the other variables, $x_1 = d_{x_1}, \ldots x_n = d_{x_n}$ with $x_i \neq x_j$ for all $i \neq j$. As this is a permutation, this corresponds to the assignment of unique variables to values. Hence, the corresponding dual all-different constraint is GAC. Finally, the channelling constraints are trivially AC. $\square$

Using these identities, we can immediately deduce, for instance, that it does not increase pruning to have both channelling constraints and primal (or dual) not-equals constraints. Not-equals constraints do not increase the amount of constraint propagation over that achieved with channelling constraints alone. As our experiments show later on, they only add overhead to the constraint solver. It is insightful to extract from these proofs the reasons why arc-consistency performs different amounts of constraint propagation in the different models. Arc-consistency deletes values in the domains of variables as follows:

**primal not-equals constraints:** if the domain of any of the primal variables is reduced to a singleton (either by constraint propagation or by assignment in a backtracking algorithm), enforcing AC on the primal not-equals constraints removes this value from all other primal variables.

**channelling constraints:** as with primal not-equals constraints; in addition, if the domain of any dual variable is reduced to a singleton, enforcing AC on the channelling constraints removes this value from all other dual variables. In particular, if a value occurs in the domain of just one other primal variable, enforcing AC on the channelling constraints ensures that no other value can be assigned to that primal variable.

**primal all-different constraint:** enforcing GAC on a primal all-different constraint will prune all the values that are removed by enforcing AC on the primal not-equals or channelling constraints. In addition, enforcing GAC is sometimes able to prune other values (e.g. if we have two primal variables with only two values between them, these values will be removed from all other primal variables).

In brief, AC on the primal not-equals constraints detects singleton variables, whilst AC on the channelling constraints detects both singleton variables *and* singleton values. GAC on a primal all-different constraint, on the other hand, determines global consistency which includes singleton variables, singleton values and many other situations.

### 5.2 Maintaining Arc-Consistency

These results can be lifted to algorithms that maintain (generalized) arc-consistency during search. Indeed, the gaps between the primal all-different and the channelling constraints, and between the channelling constraints and the primal not-equals constraints can be exponentially large. Note that not all differences in constraint tightness result in exponential reductions in search. For instance, some differences between models which are only polynomial are identified in Cheng et al. (1999). Recall that we write $X_A \Rightarrow X_B$ iff $X_A \rightarrow X_B$ and there is a problem on which algorithm $X$ visits exponentially fewer branches with $A$ than $B$. Note that $\text{GAC}_\forall$ and AC are both polynomial to enforce, so an exponential reduction in branches translates to an exponential reduction in runtime.

**Theorem 2** *On a permutation problem:*

$$MGAC_\forall \Rightarrow MAC_{\neq c \neq} \leftrightarrow MAC_{\neq c} \leftrightarrow MAC_c \Rightarrow MAC_{\neq}$$

**Proof:** We give proofs for the most important identities. Other results follow immediately from the last theorem.

To show MGAC$_\forall \Rightarrow$ MAC$_c$, consider a $(n+3)$-variable permutation problem with $x_i = \{1,\ldots,n\}$ for $i \leq n{+}1$ and $x_{n+2} = x_{n+3} = \{n{+}1, n{+}2, n{+}3\}$. Then, given a lexicographical variable ordering, MGAC$_\forall$ immediately fails, whilst MAC$_c$ takes $n!$ branches.

To show MAC$_c \Rightarrow$ MAC$_{\neq}$, consider a $(n+2)$-variable permutation problem with $x_1 = \{1,2\}$, and $x_i = \{3,\ldots,n+2\}$ for $i \geq 2$. Then, given a lexicographical variable ordering, MAC$_c$ takes 2 branches to show insolubility, whilst MAC$_{\neq}$ takes $2(n-1)!$ branches. □

### 5.3 Forward Checking

Maintaining (generalized) arc-consistency on large permutation problems can be expensive. We may therefore decide to use a cheaper local consistency property like that maintained by forward checking. For example, the Choco finite-domain toolkit in Claire uses just nFC0 on all-different constraints. The channelling constraints remain tighter than the primal not-equals constraints wrt FC.

**Theorem 3** *On a permutation problem:*

$$\begin{array}{l} nFC2_\forall \rightarrow FC_{\neq c \neq} \leftrightarrow FC_{\neq c} \leftrightarrow FC_c \rightarrow FC_{\neq} \rightarrow nFC0_\forall \\ \qquad\qquad\qquad\qquad\qquad\qquad\quad \uparrow \\ \qquad\qquad\qquad\qquad\quad nFC2_\forall \rightarrow nFC1_\forall \end{array}$$

**Proof:** Gent et al. (2000) prove FC$_{\neq} \hookrightarrow$ nFC0$_\forall$. To show strictness on permutation problems (as opposed to the more general class of decomposable constraints studied by Gent, Stergiou, and Walsh, 2000), consider a 5-variable permutation problem with $x_1 = x_2 = x_3 = x_4 = \{1,2,3\}$ and $x_5 = \{4,5\}$. Irrespective of the variable and value ordering, FC shows the problem is unsatisfiable in at most 12 branches. nFC0 by comparison takes at least 18 branches.

To show FC$_c \rightarrow$ FC$_{\neq}$, consider assigning the value $j$ to the primal variable $x_i$. FC$_{\neq}$ removes $j$ from the domain of all other primal variables. FC$_c$ instantiates the dual variable $d_j$ with the value $i$, and then removes $i$ from the domain of all other primal variables. Hence, FC$_c$ prunes all the values that FC$_{\neq}$ does. To show strictness, consider a 4-variable permutation problem with $x_1 = \{1,2\}$ and $x_2 = x_3 = x_4 = \{3,4\}$. Given a lexicographical variable and numerical value ordering, FC$_{\neq}$ shows the problem is unsatisfiable in 4 branches. FC$_c$ by comparison takes just 2 branches.

Gent et al. (2000) prove nFC1$_\forall \hookrightarrow$ FC$_{\neq}$. To show the reverse, consider assigning the value $j$ to the primal variable $x_i$. FC$_{\neq}$ removes $j$ from the domain of all primal variables except $x_i$. However, nFC1$_\forall$ also removes $j$ from the domain of all primal variables except $x_i$ since each occurs in a binary not-equals constraint with $x_i$ obtained by projecting out the all-different constraint. Hence, nFC1$_\forall \leftrightarrow$ FC$_{\neq}$.

To show nFC2$_\forall \rightarrow$ FC$_{\neq c \neq}$, consider instantiating the primal variable $x_i$ with the value $j$. FC$_{\neq c \neq}$ removes $j$ from the domain of all primal variables except $x_i$, $i$ from the domain of all dual variables except $d_j$, instantiates $d_j$ with the value $i$, and then removes $i$ from the domain of all dual variables except $d_j$. nFC2$_\forall$ also removes $j$ from the domain of all primal variables except $x_i$. The only possible difference is if one of the other dual variables, say $d_l$ has a domain wipeout. If this happens, $x_i$ has one value in its domain, $l$ that is in the domain of no other primal variable. Enforcing GAC immediately detects that $x_i$ cannot take the value $j$, and must instead take the value $k$. Hence nFC2$_\forall$ has a domain wipeout whenever FC$_{\neq c \neq}$ does. To show strictness, consider a 7-variable permutation problem with $x_1 = x_2 = x_3 = x_4 = \{1, 2, 3\}$ and $x_5 = x_6 = x_7 = \{4, 5, 6, 7\}$. Irrespective of the variable and value ordering, FC$_{\neq c \neq}$ takes at least 6 branches to show the problem is unsatisfiable. nFC2$_\forall$ by comparison takes no more than 4 branches.

Bessière et al. (1999) prove nFC2$_\forall \hookrightarrow$ nFC1$_\forall$. To show strictness on permutation problems, consider a 5-variable permutation problem with $x_1 = x_2 = x_3 = x_4 = \{1, 2, 3\}$ and $x_5 = \{4, 5\}$. Irrespective of the variable and value ordering, nFC1 shows the problem is unsatisfiable in at least 6 branches. nFC2 by comparison takes no more than 3 branches. □

### 5.4 Bounds Consistency

Another common method to reduce costs is to enforce just bounds consistency. For example, bounds consistency is used to prune a global constraint involving a sum of variables and a set of inequalities (Régin & Rueher, 2000). As a second example, some of the experiments on permutation problems performed by Smith (2000) used bounds consistency on certain of the constraints. With bounds consistency on permutation problems, we obtain a very similar ordering of the models as with AC.

**Theorem 4** *On a permutation problem:*

$$\begin{array}{ccccccccc} BC_\forall & \rightarrow & BC_{\neq c \neq} & \leftrightarrow & BC_{\neq c} & \leftrightarrow & BC_c & \rightarrow & BC_{\neq} \\ & & & & & & \uparrow & & \\ & & & & & & AC_{\neq} & & \end{array}$$

**Proof:** To show BC$_c \rightarrow$ BC$_{\neq}$, consider a permutation problem which is BC$_c$ but one of the primal not-equals constraints is not BC. Then, it would involve two variables, $x_i$ and $x_j$ both with identical interval domains, $[k, k]$. Enforcing BC on the channelling constraint between $x_i$ and $d_k$ would reduce $d_k$ to the domain $[i, i]$. Enforcing BC on the channelling constraint between $x_j$ and $d_k$ would then cause a domain wipeout. But this contradicts the channelling constraints being BC. Hence, all the primal not-equals constraints must be BC. To show strictness. consider a 3-variable permutation problem with $x_1 = x_2 = [1, 2]$ and $x_3 = [1, 3]$. This is BC$_{\neq}$ but not BC$_c$.

To show BC$_\forall \rightarrow$ BC$_{\neq c \neq}$, consider a permutation problem which is BC$_\forall$. Suppose we assign a boundary value $j$ to a primal variable, $x_i$ (or equivalently, a boundary value $i$ to a dual variable, $d_j$). As the all-different constraint is BC, this can be extended to all the other primal variables using each of the values once. This gives us a consistent assignment for any other primal or dual variable. Hence, it is BC$_{\neq c \neq}$. To show strictness, consider a 5-variable permutation problem with $x_1 = x_2 = x_3 = [1, 2]$ and $x_4 = x_5 = [3, 5]$. This is BC$_{\neq c \neq}$ but not BC$_\forall$.

To show AC$_{\neq} \rightarrow$ BC$_c$, consider a permutation problem which is BC$_c$ but not AC$_{\neq}$. Then there must be one constraint, $x_i \neq x_j$, with $x_i$ and $x_j$ having the same singleton domain, $\{k\}$. But, if this is the case, enforcing BC on the channelling constraints between $x_i$ and $d_k$ and between $x_j$ and $d_k$ would prove that the problem is unsatisfiable. Hence, it is AC$_{\neq}$. To show strictness, consider a 3-variable permutation problem with $x_1 = x_2 = [1, 2]$ and $x_3 = [1, 3]$. This is AC$_{\neq}$ but not BC$_c$. □

### 5.5 Restricted Path Consistency

Debruyne and Bessière (1997) have shown that RPC is a promising filtering technique above AC. It prunes many of the PIC values at little extra cost to AC. Surprisingly, channelling constraints are incomparable to the primal not-equals constraints wrt RPC. Channelling constraints can increase the amount of propagation (for example, when a dual variable has only one value left in its domain). However, RPC is hindered by the bipartite constraint graph between primal and dual variables. Additional not-equals constraints on primal and/or dual variables can therefore help propagation.

**Theorem 5** *On a permutation problem;*

$GAC_{\forall} \rightarrow RPC_{\neq c \neq} \rightarrow RPC_{\neq c} \rightarrow RPC_c \otimes RPC_{\neq} \otimes AC_c$

**Proof:** To show $\mathrm{RPC}_c \otimes \mathrm{RPC}_{\neq}$, consider a 4-variable permutation problem with $x_1 = x_2 = x_3 = \{1,2,3\}$ and $x_4 = \{1,2,3,4\}$. This is $\mathrm{RPC}_{\neq}$ but not $\mathrm{RPC}_c$. For the reverse direction, consider a 5-variable permutation problem with $x_1 = x_2 = x_3 = \{1,2\}$ and $x_4 = x_5 = \{3,4,5\}$. This is $\mathrm{RPC}_c$ but not $\mathrm{RPC}_{\neq}$.

To show $\mathrm{RPC}_{\neq c} \rightarrow \mathrm{RPC}_c$, consider again the last example. This is $\mathrm{RPC}_c$ but not $\mathrm{RPC}_{\neq c}$.

To show $\mathrm{RPC}_{\neq c \neq} \rightarrow \mathrm{RPC}_{\neq c}$, consider a 6-variable permutation problem with $x_1 = x_2 = \{1,2,3,4,5,6\}$ and $x_3 = x_4 = x_5 = x_6 = \{4,5,6\}$. This is $\mathrm{RPC}_{\neq c}$ but not $\mathrm{RPC}_{\neq c \neq}$.

To show $\mathrm{GAC}_{\forall} \rightarrow \mathrm{RPC}_{\neq c \neq}$, consider a permutation problem which is $\mathrm{GAC}_{\forall}$. Suppose we assign a value $j$ to a primal variable, $x_i$ (or equivalently, a value $i$ to a dual variable, $d_j$). As the all-different constraint is GAC, this can be extended to all the other primal variables using up all the other values. This gives us a consistent assignment for any two other primal or dual variables. Hence, the problem is $\mathrm{PIC}_{\neq c \neq}$ and thus $\mathrm{RPC}_{\neq c \neq}$. To show strictness, consider a 7-variable permutation problem with $x_1 = x_2 = x_3 = x_4 = \{1,2,3\}$ and $x_5 = x_6 = x_7 = \{4,5,6,7\}$. This is $\mathrm{RPC}_{\neq c \neq}$ but not $\mathrm{GAC}_{\forall}$.

To show $\mathrm{AC}_c \otimes \mathrm{RPC}_{\neq}$, consider a 4-variable permutation problem with $x_1 = x_2 = x_3 = \{1,2,3\}$ and $x_4 = \{1,2,3,4\}$. This is $\mathrm{RPC}_{\neq}$ but not $\mathrm{AC}_c$. For the reverse direction, consider a 5-variable permutation problem with $x_1 = x_2 = x_3 = \{1,2\}$ and $x_4 = x_5 = \{3,4,5\}$. This is $\mathrm{AC}_c$ but not $\mathrm{RPC}_{\neq}$. □

### 5.6 Path Inverse Consistency

The incomparability of channelling constraints and primal not-equals constraints remains when we move up the local consistency hierarchy from RPC to PIC.

**Theorem 6** *On a permutation problem:*

$GAC_{\forall} \rightarrow PIC_{\neq c \neq} \rightarrow PIC_{\neq c} \rightarrow PIC_c \otimes PIC_{\neq} \otimes AC_c$

**Proof:** To show $\mathrm{PIC}_c \otimes \mathrm{PIC}_{\neq}$, consider a 4-variable permutation problem with $x_1 = x_2 = x_3 = \{1,2,3\}$ and $x_4 = \{1,2,3,4\}$. This is $\mathrm{PIC}_{\neq}$ but not $\mathrm{PIC}_c$. Enforcing PIC on the channelling constraints reduces $x_4$ to the singleton domain $\{4\}$. For the reverse direction, consider a 5-variable permutation problem with $x_1 = x_2 = x_3 = \{1,2\}$ and $x_4 = x_5 = \{3,4,5\}$. This is $\mathrm{PIC}_c$ but not $\mathrm{PIC}_{\neq}$.

To show $\text{PIC}_{\neq c} \to \text{PIC}_c$, consider a 5-variable permutation problem with $x_1 = x_2 = x_3 = \{1,2\}$ and $x_4 = x_5 = \{3,4,5\}$. This is $\text{PIC}_c$ but not $\text{PIC}_{\neq c}$.

To show $\text{PIC}_{\neq c\neq} \to \text{PIC}_{\neq c}$, consider a 6-variable permutation problem with $x_1 = x_2 = \{1,2,3,4,5,6\}$ and $x_3 = x_4 = x_5 = x_6 = \{4,5,6\}$. This is $\text{PIC}_{\neq c}$ but not $\text{PIC}_{\neq c\neq}$.

To show $\text{GAC}_\forall \to \text{PIC}_{\neq c\neq}$, consider a permutation problem in which the all-different constraint is GAC. Suppose we assign a value $j$ to a primal variable, $x_i$ (or equivalently, a value $i$ to a dual variable, $d_j$). As the all-different constraint is GAC, this can be extended to all the other primal variables using up all the other values. This gives us a consistent assignment for any two other primal or dual variables. Hence, the not-equals and channelling constraints are PIC. To show strictness, consider a 7-variable permutation problem with $x_1 = x_2 = x_3 = x_4 = \{1,2,3\}$ and $x_5 = x_6 = x_7 = \{4,5,6,7\}$. This is $\text{PIC}_{\neq c\neq}$ but not $\text{GAC}_\forall$.

To show $\text{PIC}_{\neq} \otimes \text{AC}_c$, consider a 4-variable permutation problem with $x_1 = x_2 = x_3 = \{1,2,3\}$ and $x_4 = \{1,2,3,4\}$. This is $\text{PIC}_{\neq}$ but not $\text{AC}_c$. Enforcing AC on the channelling constraints reduces $x_4$ to the singleton domain $\{4\}$. For the reverse direction, consider a 5-variable permutation problem with $x_1 = x_2 = x_3 = \{1,2\}$ and $x_4 = x_5 = \{3,4,5\}$. This is $\text{AC}_c$ but not $\text{PIC}_{\neq}$. □

### 5.7 Singleton Arc-Consistency

Debruyne and Bessière (1997) also showed that SAC is a promising filtering technique above both AC, RPC and PIC, pruning many values for its CPU time. Prosser et al. (2000) reported promising experimental results with SAC on quasigroup problems, a multiple permutation problem. Interestingly, as with AC (but unlike RPC and PIC which lie between AC and SAC), channelling constraints are tighter than the primal not-equals constraints wrt SAC.

**Theorem 7** *On a permutation problem:*

$$GAC_\forall \to SAC_{\neq c\neq} \leftrightarrow SAC_{\neq c} \leftrightarrow SAC_c \to SAC_{\neq} \otimes AC_c$$

**Proof:** To show $\text{SAC}_c \to \text{SAC}_{\neq}$, consider a permutation problem that is $\text{SAC}_c$ and any instantiation for a primal variable $x_i$. Suppose that the primal not-equals model of the resulting problem cannot be made AC. Then there must exist two other primal variables, say $x_j$ and $x_k$ which have at most one other value. Consider the dual variable associated with this value. Then under this instantiation of the primal variable $x_i$, enforcing AC on the channelling constraint between the primal variable $x_i$ and the dual variable, and between the dual variable and $x_j$ and $x_k$ results in a domain wipeout on the dual variable. Hence the problem is not $\text{SAC}_c$. This is a contradiction. The primal not-equals model can therefore be made AC following the instantiation of $x_i$. That is, the problem is $\text{SAC}_{\neq}$. To show strictness, consider a 5-variable permutation problem with domain $x_1 = x_2 = x_3 = x_4 = \{0,1,2\}$ and $x_5 = \{3,4\}$. This is $\text{SAC}_{\neq}$ but not $\text{SAC}_c$.

To show $\text{GAC}_\forall \to \text{SAC}_c$, consider a permutation problem that is $\text{GAC}_\forall$. Consider any instantiation for a primal variable. This can be consistently extended to all variables in the primal model. But this means that it can be consistently extended to all variables in the primal and dual model, satisfying any (combination of) permutation or channelling constraints. As the channelling constraints are satisfiable, they can be made AC. Consider any instantiation for a dual variable. By a similar argument, taking the appropriate instantiation for the associated primal variable, the resulting problem can be made AC. Hence, given any instantiation for a primal or dual variable, the channelling constraints can be made AC. That is, the problem is SAC$_c$, To show strictness, consider a 7-variable permutation problem with $x_1 = x_2 = x_3 = x_4 = \{0, 1, 2\}$ and $x_5 = x_6 = x_7 = \{3, 4, 5, 6\}$. This SAC$_c$ but is not GAC$_\forall$.

To show SAC$_{\neq}$ $\otimes$ AC$_c$, consider a four variable permutation problem in which $x_1$ to $x_3$ have the $\{1, 2, 3\}$ and $x_4$ has the domain $\{0, 1, 2, 3\}$. This is SAC$_{\neq}$ but not AC$_c$. For the reverse, consider a 4-variable permutation problem with $x_1 = x_2 = \{0, 1\}$ and $x_3 = x_4 = \{0, 2, 3\}$. This is AC$_c$ but not SAC$_{\neq}$. $\square$

### 5.8 Strong Path-Consistency

Adding primal or dual not-equals constraints to channelling constraints does not help AC or SAC. The following result shows that their addition does not help higher levels of local consistency like strong path-consistency (ACPC).

**Theorem 8** *On a permutation problem:*

$$GAC_\forall \otimes ACPC_{\neq c \neq} \leftrightarrow ACPC_{\neq c} \leftrightarrow ACPC_c \rightarrow ACPC_{\neq} \otimes AC_c$$

**Proof:** To show ACPC$_c$ $\rightarrow$ ACPC$_{\neq}$, consider some channelling constraints that are ACPC. Now AC$_c$ $\rightarrow$ AC$_{\neq}$, so we just need to show PC$_c$ $\rightarrow$ PC$_{\neq}$. Consider a consistent pair of values, $l$ and $m$ for a pair of primal variables, $x_i$ and $x_j$. Take any third primal variable, $x_k$. As the constraint between $d_l$, $d_m$ and $x_k$ is PC, we can find a value for $x_k$ consistent with the channelling constraints. But this also satisfies the not-equals constraint between primal variables. Hence, the problem is PC$_{\neq}$. To show strictness, consider a 4-variable permutation problem with $x_1 = x_2 = x_3 = x_4 = \{1, 2, 3\}$. This is ACPC$_{\neq}$ but not ACPC$_c$.

To show ACPC$_{\neq c \neq}$ $\leftrightarrow$ ACPC$_{\neq c}$ $\leftrightarrow$ ACPC$_c$, we recall that AC$_{\neq c}$ $\leftrightarrow$ AC$_{\neq c}$ $\leftrightarrow$ AC$_c$. Hence we need just show that PC$_{\neq c}$ $\leftrightarrow$ PC$_{\neq c}$ $\leftrightarrow$ PC$_c$. Consider a permutation problem. Enforcing PC on the channelling constraints alone infers both the primal and the dual not-equals constraints. Hence, PC$_{\neq c}$ $\leftrightarrow$ PC$_{\neq c}$ $\leftrightarrow$ PC$_c$.

To show GAC$_\forall$ $\otimes$ ACPC$_{\neq c \neq}$, consider a 6-variable permutation problem with $x_1 = x_2 = x_3 = x_4 = \{1, 2, 3\}$, and $x_5 = x_6 = \{4, 5, 6\}$. This is ACPC$_{\neq c \neq}$ but not GAC$_\forall$. For the reverse direction, consider a 3-variable permutation problem with $x_1 = x_2 = x_3 = \{1, 2, 3\}$, and the additional binary constraint $even(x_1 + x_3)$. Enforcing GAC$_\forall$ prunes the domains to $x_1 = x_3 = \{1, 3\}$, and $x_2 = \{2\}$. However, these domains are not ACPC$_{\neq c \neq}$. Enforcing ACPC tightens the constraint between $x_1$ and $x_3$ from not-equals to $x_1 = 1, x_3 = 3$ or $x_1 = 3, x_3 = 1$.

To show ACPC$_{\neq}$ $\otimes$ AC$_c$, consider a 5-variable permutation problem with $x_1 = x_2 = x_3 = \{1, 2\}$, and $x_4 = x_5 = \{3, 4, 5\}$. This is AC$_c$ but not ACPC$_{\neq}$. For the reverse direction, consider again the 4-variable permutation problem with $x_1 = x_2 = x_3 = x_4 = \{1, 2, 3\}$. This is ACPC$_{\neq}$ but not AC$_c$. $\square$

### 5.9 Multiple Permutation Problems

These results extend to multiple permutation problems under a simple restriction that the problem is *triangle preserving* (Stergiou & Walsh, 1999). That is, any triple of variables which are all-different must occur together in at least one permutation. For example, the three constraints all-diff($x_1, x_2, x_4$), all-diff($x_1, x_3, x_5$), and all-diff($x_2, x_3, x_6$) are not triangle preserving as $x_1$, $x_2$ and $x_3$ are all-different but are not in the same constraint. The following theorem collects together and generalizes many of the previous results.

**Theorem 9** *On a multiple permutation problem:*

$$\begin{array}{llllllllll}
GAC_\forall & \otimes ACPC_{\neq c \neq} & \leftrightarrow & ACPC_{\neq c} & \leftrightarrow & ACPC_c & \rightarrow & ACPC_{\neq} & \otimes & AC_c \\
 & \downarrow & & \downarrow & & \downarrow & & \downarrow & & \\
GAC_\forall & \rightarrow SAC_{\neq c \neq} & \leftrightarrow & SAC_{\neq c} & \leftrightarrow & SAC_c & \rightarrow & SAC_{\neq} & \otimes & AC_c \\
 & \downarrow & & \downarrow & & \downarrow & & \downarrow & & \\
GAC_\forall & \rightarrow PIC_{\neq c \neq} & \rightarrow & PIC_{\neq c} & \rightarrow & PIC_c & \otimes & PIC_{\neq} & \otimes & AC_c \\
 & \downarrow & & \downarrow & & \downarrow & & \downarrow & & \\
GAC_\forall & \rightarrow RPC_{\neq c \neq} & \rightarrow & RPC_{\neq c} & \rightarrow & RPC_c & \otimes & RPC_{\neq} & \otimes & AC_c \\
 & \downarrow & & \downarrow & & \downarrow & & \downarrow & & \\
GAC_\forall & \rightarrow AC_{\neq c \neq} & \leftrightarrow & AC_{\neq c} & \leftrightarrow & AC_c & \rightarrow & AC_{\neq} & \rightarrow & BC_c \\
\downarrow & \downarrow & & \downarrow & & \downarrow & & \downarrow & & \\
BC_\forall & \rightarrow BC_{\neq c \neq} & \leftrightarrow & BC_{\neq c} & \leftrightarrow & BC_c & \rightarrow & BC_{\neq} & &
\end{array}$$

**Proof:** The proofs lift in a straightforward manner from the single permutation case. Local consistencies like ACPC, SAC, PIC and RPC consider triples of variables. If these are linked together, we use the fact that the problem is triangle preserving and a permutation is therefore defined over them. If these are not linked together, we can decompose the argument into AC on pairs of variables. Without triangle preservation, $GAC_\forall$, may only achieve as high a level of consistency as $AC_{\neq}$. For example, consider again the non-triangle preserving constraints in the last paragraph. If $x_1 = x_2 = x_3 = \{1, 2\}$ and $x_4 = x_5 = x_6 = \{1, 2, 3\}$ then the problem is $GAC_\forall$, but it is not $RPC_{\neq}$, and hence neither $PIC_{\neq}$, $SAC_{\neq}$ nor $ACPC_{\neq}$. □

## 6. SAT Models

Another solution strategy is to encode permutation problems into SAT and use a fast Davis-Putnam (DP) or local search procedure. For example, Bejar and Manya (2000) report promising results for propositional encodings of round robin problems, which include permutation constraints. We consider here just "direct" encodings into SAT as these have been used most commonly in the past (Walsh, 2000). An alternative and promising encoding of CSPs into SAT is the "support encoding". Recently, Gent (2002) has shown that unit propagation in the support encoding is equivalent to enforcing arc-consistency in the original CSP, and this can be achieved in asymptotically optimal time. To compare the support encodings of the different models of a permutation problem, we simply need therefore to look at our results on arc-consistency. With the direct encoding, unit propagation enforces a level of local consistency less than arc-consistency. Indeed, the level of consistency is often identical to that achieved by the forward checking algorithm.

In the direct encoding of a CSP into SAT, we have a Boolean variable $X_{ij}$ which is *true* iff the primal variable $x_i$ takes the value $j$. In the primal SAT model, there are $n$ clauses to ensure that each primal variable takes at least one value, $O(n^3)$ clauses to ensure that no primal variable gets two values, and $O(n^3)$ clauses to ensure that no two primal variables take the same value. Interestingly the channelling SAT model has the same number of Boolean variables as the primal SAT model (as we can use $X_{ij}$ to represent both the $j$th value of the primal variable $x_i$ *and* the $i$th value for the dual variable $d_j$), and just $n$ additional clauses to ensure each dual variable takes a value. The $O(n^3)$ clauses to ensure that no dual variable gets two values are equivalent to the clauses that ensure no two primal variables get the same value. The following results show that MAC is tighter than DP, and DP is equivalent to FC on these different models. In what follows, we assume that the FC algorithm uses a fail first heuristic that instantiates variables with single values left in their domains before variables with a choice of values (Haralick & Elliot, 1980).

**Theorem 10** *On a permutation problem:*

$$\begin{array}{ccccccccc}
MGAC_{\forall} & \rightarrow & MAC_{\neq c \neq} & \leftrightarrow & MAC_{\neq c} & \leftrightarrow & MAC_{c} & \rightarrow & MAC_{\neq} \\
 & & \downarrow & & \downarrow & & \downarrow & & \downarrow \\
MGAC_{\forall} & \rightarrow & DP_{\neq c \neq} & \leftrightarrow & DP_{\neq c} & \leftrightarrow & DP_{c} & \rightarrow & DP_{\neq} \\
 & & \updownarrow & & \updownarrow & & \updownarrow & & \updownarrow \\
MGAC_{\forall} & \rightarrow & FC_{\neq c \neq} & \leftrightarrow & FC_{\neq c} & \leftrightarrow & FC_{c} & \rightarrow & FC_{\neq}
\end{array}$$

**Proof:** $\text{DP}_{\neq} \leftrightarrow \text{FC}_{\neq}$ is a special case of Theorem 14 (Walsh, 2000), whilst $\text{MAC}_{\neq} \rightarrow \text{FC}_{\neq}$ is a special case of Theorem 15.

To show $\text{DP}_c \leftrightarrow \text{FC}_c$ suppose unit propagation sets a literal $l$. There are four cases. In the first case, a clause of the form $X_{i1} \vee \ldots \vee X_{in}$ has been reduced to an unit. That is, we have one value left for a primal variable. The fail first heuristic in FC picks this last value to instantiate. In the second case, a clause of the form $\neg X_{ij} \vee \neg X_{ik}$ for $j \neq k$ has been reduced to an unit. This ensures that no primal variable gets two values. The FC algorithm trivially never tries two simultaneous values for a primal variable. In the third case, a clause of the form $\neg X_{ij} \vee \neg X_{kj}$ for $i \neq k$ has been reduced to an unit. This ensures that no dual variable gets two values. Again, the FC algorithm trivially never tries two simultaneous values for a dual variable. In the fourth case, $X_{1j} \vee \ldots \vee X_{nj}$ has been reduced to an unit. That is, we have one value left for a dual variable. A fail first heuristic in FC picks this last value to instantiate. Hence, given a suitable branching heuristic, the FC algorithm tracks the DP algorithm. To show the reverse, suppose forward checking removes a value. There are two cases. In the first case, the value $i$ is removed from a dual variable $d_j$ due to some channelling constraint. This means that there is a primal variable $x_k$ which has been set to some value $l \neq j$. Unit propagation on $\neg X_{kl} \vee \neg X_{kj}$ sets $X_{kj}$ to false, and then on $\neg X_{ij} \vee \neg X_{kj}$ sets $X_{ij}$ to false as required. In the second case, the value $i$ is removed from a dual variable $d_j$, again due to a channelling constraint. The proof is now dual to the first case.

To show $\text{MAC}_c \rightarrow \text{DP}_c$, we use the fact that MAC dominates FC and $\text{FC}_c \leftrightarrow \text{DP}_c$. To show strictness, consider a 3-variable permutation problem with additional binary constraints that rule out the same value for all 3 primal variables. Enforcing AC on the channelling constraints causes a domain wipeout on the dual variable associated with this value. As there are no unit clauses, DP does not immediately solve the problem.

To show $DP_c \rightarrow DP_{\neq}$, we note that the channelling SAT model contains more clauses. Hence, it dominates the primal SAT model. To show strictness, consider a four variable permutation problem with three additional binary constraints that if $x_1 = 1$ then $x_2 = 2$, $x_3 = 2$ and $x_4 = 2$ are all ruled out. Consider branching on $x_1 = 1$. Unit propagation on both models sets $X_{12}$, $X_{22}$, $X_{32}$, $X_{42}$, $X_{21}$, $X_{31}$ and $X_{41}$ to false. On the channelling SAT model, unit propagation against the clause $X_{12} \vee X_{22} \vee X_{32} \vee X_{42}$ then generates an empty clause. By comparison, unit propagation on the primal SAT model does no more work. □

## 7. Asymptotic Comparison

The previous results tell us nothing about the relative cost of achieving these local consistencies. Asymptotic analysis adds detail to the results. We can achieve $GAC_{\forall}$ in $O(n^4)$ time (Régin, 1994). AC on binary constraints can be achieved in $O(ed^2)$ where $e$ is the number of constraints and $d$ is their domain size. As there are $O(n^2)$ channelling constraints, $AC_c$ naively takes $O(n^4)$ time. However, by taking advantage of the functional nature of channelling constraints, we can reduce this to $O(n^3)$ using the AC-5 algorithm (Hentenryck, Deville, & Teng, 1992). $AC_{\neq}$ also naively takes $O(n^4)$ time as there are $O(n^2)$ binary not-equals constraints. However, we can take advantage of the special nature of a binary not-equals constraint to reduce this to $O(n^2)$ as each not-equals constraint needs to be made AC just once. We have proved that $GAC_{\forall} \rightarrow AC_c \rightarrow AC_{\neq}$ and greater pruning power is reflected in higher worst case complexity ($O(n^4)$, $O(n^3)$, $O(n^2)$ respectively). Thus we still need to run experiments to see if the additional pruning outweighs the potentially higher cost.

## 8. Experimental Comparison

We ran a wide variety of experiments to explore the significance of these theoretical and asymptotic differences. For example, even though binary not-equals constraints do less pruning than the channelling constraints, they might still speed up search by pruning quicker. We limit the first set of experiments to a static variable and value ordering as we wish to confirm the theoretical results, and these are limited either to static orderings or to a restricted class of dynamic variable and value orderings in which we make "equivalent" branching decisions in the different search trees (Bacchus et al., 2002).

As explained before, many constraint toolkits support channelling with efficient global constraints. For example, ILOG Solver has the `IlcInverse` constraint, and the Sicstus finite domain constraint library has the `assignment` predicate. Both perform a level of pruning which appears to be equivalent to enforcing AC on the explicit channelling constraints. We therefore compared this in our experiments to AC on the binary not-equals constraints and GAC on the all-different constraint. All the models are implemented in Solver 5.300, and are available via CSPLib. We lexicographically order the variables and assign the values in numerical order. We therefore only branch on primal variables. As we observe very similar results on a range of permutation problems, we only show here results for Langford's problem.

| | | L(3,9) | | L(3,10) | |
|---|---|---|---|---|---|
| model | heuristic | fails | sec. | fails | sec. |
| $\forall$ | static | **12** | **0.001** | **42** | **0.003** |
| $c$ | static | **12** | 0.003 | 43 | 0.005 |
| $\neq$ | static | 25 | **0.001** | 82 | 0.011 |
| $\neq c$ | static | **12** | 0.005 | 43 | 0.013 |
| $c\neq$ | static | **12** | **0.001** | 43 | 0.011 |
| $\forall c$ | static | **12** | **0.001** | **42** | 0.009 |
| $c\forall$ | static | **12** | 0.003 | **42** | 0.009 |
| $\neq c\neq$ | static | **12** | 0.005 | 43 | 0.015 |
| $\forall c\neq$ | static | **12** | 0.005 | **42** | 0.011 |
| $\neq c\forall$ | static | **12** | 0.007 | **42** | 0.013 |
| $\forall c\forall$ | static | **12** | 0.003 | **42** | 0.009 |

Table 3: Number of backtracks (fails) and running time to find the first solution to two instances of Langford's problem. Runtimes are for ILOG Solver 5.300 on a 1200MHz, Pentium III processor, and 512 MB of RAM.

| | | L(3,9) | | L(3,10) | | L(3,11) | | L(3,12) | |
|---|---|---|---|---|---|---|---|---|---|
| model | heuristic | fails | sec. | fails | sec. | fails | sec. | fails | sec. |
| $\forall$ | static | **2006** | **0.22** | **10051** | **1.13** | **49118** | **5.86** | **279468** | **35.36** |
| $c$ | static | 2282 | 0.28 | 11336 | 1.45 | 56234 | 7.41 | 312926 | 41.89 |
| $\neq$ | static | 6062 | 0.59 | 29018 | 3.15 | 167624 | 20.59 | 949878 | 131.04 |
| $\neq c$ | static | 2282 | 0.41 | 11336 | 2.26 | 56234 | 11.91 | 312926 | 72.85 |
| $c\neq$ | static | 2282 | 0.41 | 11336 | 2.25 | 56234 | 11.94 | 312926 | 72.2 |
| $\forall c$ | static | **2006** | 0.32 | **10051** | 1.72 | **49118** | 8.61 | **279468** | 50.53 |
| $c\forall$ | static | **2006** | 0.33 | **10051** | 1.76 | **49118** | 8.77 | **279468** | 51.41 |
| $\neq c\neq$ | static | 2282 | 0.53 | 11336 | 3.21 | 56234 | 18.21 | 312926 | 114.44 |
| $\forall c\neq$ | static | **2006** | 0.43 | **10051** | 2.38 | **49118** | 12.32 | **279468** | 76.77 |
| $\neq c\forall$ | static | **2006** | 0.66 | **10051** | 2.49 | **49118** | 12.92 | **279468** | 78.95 |
| $\forall c\forall$ | static | **2006** | 0.39 | **10051** | 2.09 | **49118** | 10.56 | **279468** | 62.49 |

Table 4: Number of backtracks (fails) and running time to find all solutions, or prove that there are no solutions, to four instances of Langford's problem. Runtimes are for ILOG Solver 5.300 on 1200MHz, Pentium III processor, and 512 MB of RAM.

In Table 3, we compare the various models of a permutation when finding the first solution to two instances of Langford's problem. In Table 4, we compare the same models when finding all solutions or proving that there are no solutions, for four instances of Langford's problem. Only L(3,9) and L(3,10) in this table have any solutions. The experimental results confirm our theoretical findings. First, enforcing GAC on an all-different constraint does the most pruning, whilst enforcing AC on the binary not-equals constraints does the least, and enforcing AC on the channelling constraints is in between. Runtimes are similarly ordered. Second, adding the primal or dual binary not-equals constraints to the channelling constraints does not bring any more pruning, and merely adds overhead to the runtime. Third, adding extra constraints to the primal or dual all-different constraint achieves the same amount of pruning as the all-different constraint on its own, and again just adds overhead to the runtime.

## 9. Dynamic Variable And Value Ordering

The experimental results in the last section might seem to have settled the matter of how to model permutation problems. Enforcing GAC on a single all-different constraint always gave the smallest search trees and runtimes. However, this ignores a significant potential advantage of channelling into a dual model. Dynamic variable and value ordering heuristics may be able to exploit the primal and dual viewpoints of a permutation to make better decisions. This is not a topic that can be easily addressed theoretically. However, the experimental results given in this section show that variable and value ordering heuristics can profit greatly from multiple viewpoints.

A variable ordering heuristic like smallest domain is usually justified in terms of a fail-first principle: we have to pick eventually all the variables, so it is wise to choose one that is hard to assign, giving us hopefully much constraint propagation and a small search tree. A value ordering heuristic like maximum promise (Geelen, 1992) is usually justified in terms of a succeed-first principle: we pick a value likely to lead to a solution, so reducing the risk of backtracking and trying one of the alternative values. In a permutation problem, we can branch on the primal or the dual variables or on both. We shall show here that fail-first on one viewpoint is compatible with succeed-first on the dual. To do so, we consider the following heuristics.

**Smallest domain, SD(p+d)** : choose the primal or the dual variable with the smallest domain, and choose the values in numeric order.

**Primal smallest domain, SD(p)** : choose the primal variable with the smallest domain, and choose the values in numeric order.

**Dual smallest domain, SD(d)** : choose the dual variable with the smallest domain, and choose the values in numeric order.

**Double smallest domain, $\mathbf{SD^2(p+d)}$** : choose the primal/dual variable with the smallest domain, and choose the value whose dual/primal variable has the smallest domain.

**Primal double smallest domain, $\mathbf{SD^2(p)}$** : choose the primal variable with the smallest domain, and choose the value whose dual variable has the smallest domain.

**Dual double smallest domain, $SD^2(d)$** : choose the dual variable with the smallest domain, and choose the value whose primal variable has the smallest domain.

The smallest domain heuristic on the dual has been used as a value ordering heuristic in a number of experimental studies (Jourdan, 1995; Cheng et al., 1999; Smith, 2000). The following argument shows that the double smallest domain heuristics are compatible with the fail first principle for variable ordering and succeed first for value ordering. Suppose we assign the primal value $j$ to the primal variable $x_i$ (an analogous argument can be given if we branch on a dual variable). Constraint propagation will prune the primal value $j$ from the other primal variables, and the dual value $i$ from the other dual variables. Constraint propagation may do more than this if we have an all-different constraint or channelling constraints. However, to a first approximation, this is a reasonable starting point. The succeed first value ordering heuristic computes the "promise" of the different values by multiplying together the domain sizes of the uninstantiated variables (Geelen, 1992). Any term in this product is unchanged if $j$ or $i$, depending on whether this is a primal or dual variable, does not occur in the domain and is reduced by 1 if $j$ or $i$ occurs. The product is likely to be maximized by ensuring we reduce as few terms as possible. That is, by ensuring $j$ and $i$ occur in as few domains as possible. That is $d_j$ and $x_i$ have the smallest domains possible. Hence double smallest domain will branch on the variable with smallest domain and tend to assign it the value with most promise.

We now compare these heuristics in an extensive set of experiments. The hypothesis we wish to test is that branching heuristics can profit from multiple viewpoints. We use the following collection of permutation problems in addition to Langford's problem:

**Quasigroup existence problem:** An order $m$ quasigroup is a Latin square of size $m$, that is, an $m \times m$ multiplication table in which each element occurs in every row and every column. Quasigroup existence problems determine the existence or non-existence of quasigroups of a given size with additional properties:

- QG3($m$): denotes quasigroups of order $m$ for which $(a * b) * (b * a) = a$.
- QG4($m$): denotes quasigroups of order $m$ for which $(b * a) * (a * b) = a$.

We additionally demand that the quasigroup is idempotent, i.e. $a * a = a$ for every element $a$. The problem is **prob003** in CSPLib.

**Golomb rulers problem:** A Golomb ruler consists of $n$ marks arranged along a ruler of length $m$ such that the distances between any pair of marks form a permutation. The problem is **prob006** at CSPLib. In our experiments we specify the known optimal length and find all optimal solutions.

**Sport scheduling problem:** The problem consists of scheduling games between $n$ teams over $n - 1$ weeks when $n$ is even ($n$ weeks when $n$ is odd). Each week is divided into $n/2$ periods when $n$ is even ($(n - 1)/2$ when $n$ is odd). Each game is composed of two slots, "home" and "away", where one team plays home and the other team plays away. The objective is to schedule a game for each period of every week such that: every team plays against every other team; a team plays exactly once a week when we have an even number of teams, and at most once a week when we have an odd number of weeks; and a team plays at most twice in the same period over the course of the season. The problem is **prob026** in CSPLib.

**Magic squares problem:** An order $n$ magic square is an $n$ by $n$ matrix containing the numbers 1 to $n^2$, with the sum of each row, column, and diagonal being equal. The problem is **prob019** in CSPLib.

### 9.1 Langford's Problem

| | | L(3,12) | | L(3,13) | | L(3,14) | | L(3,15) | |
|---|---|---|---|---|---|---|---|---|---|
| model | heuristic | fails | sec. | fails | sec. | fails | sec. | fails | sec. |
| $\neq$ | SD(p) | 62016 | 10.27 | 300800 | 53.72 | 1368322 | 272.03 | 7515260 | 1601.00 |
| $\forall$ | SD(p) | 20795 | 3.59 | 93076 | 16.95 | 405519 | 78.18 | 2072534 | 414.71 |
| $c$ | SD(p+d) | 11683 | **2.16** | 45271 | **8.66** | 184745 | **36.46** | 846851 | **171.97** |
| $c$ | SD(p) | 21148 | 3.68 | 94795 | 16.84 | 412882 | 74.99 | 2112477 | 389.69 |
| $c$ | SD(d) | 15214 | 2.64 | 59954 | 10.73 | 249852 | 46.39 | 1144168 | 221.01 |
| $c$ | SD$^2$(p+d) | 11683 | 2.2 | 45271 | 9.04 | 184745 | 38.32 | 846851 | 180.00 |
| $c$ | SD$^2$(p) | 20855 | 3.89 | 93237 | 17.07 | 406546 | 75.38 | 2077692 | 393.21 |
| $c$ | SD$^2$(d) | 14314 | 2.62 | 56413 | 10.61 | 234770 | 45.68 | 1076352 | 213.51 |
| $\forall c$ | SD(p+d) | **11449** | 2.84 | **44253** | 11.47 | **180611** | 48.71 | **827564** | 231.80 |
| $\forall c$ | SD(p) | 20795 | 4.93 | 93076 | 22.61 | 405519 | 102.45 | 2072534 | 537.14 |
| $\forall c$ | SD(d) | 14459 | 3.44 | 56701 | 13.94 | 234790 | 60.13 | 1069249 | 282.42 |
| $\forall c$ | SD$^2$(p+d) | 11451 | 2.91 | 44254 | 11.72 | 180631 | 49.71 | 827605 | 235.56 |
| $\forall c$ | SD$^2$(p) | 20488 | 4.98 | 91513 | 22.86 | 399092 | 103.09 | 2037159 | 540.04 |
| $\forall c$ | SD$^2$(d) | 13639 | 3.38 | 53483 | 13.78 | 221307 | 59.33 | 1009250 | 278.32 |

Table 5: Number of backtracks (fails) and running time to find all solutions, or prove that there are no solutions, to four instances of Langford problem. Runtimes are for ILOG Solver 5.300 on 1200MHz, Pentium III processor, and 512 MB of RAM.

The results are given in Table 5. We make a number of observations. Enforcing AC on the primal not-equals model ("$\neq$") gives the worst results (as it does in almost all the subsequent problem domains). We will not therefore discuss it further. The best runtimes are obtained with the $c$ model, heuristic SD(p+d), i.e. from enforcing a permutation by the channelling constraints alone and choosing the variable with smallest domain, whether primal or dual. Using just the primal or just the dual variables as decision variables tends to increase runtimes. The branching heuristic does indeed profit from the multiple viewpoints. Note that the $\forall$ model is no longer the best strategy, in terms of either failures or runtimes, as it was in Table 4. This is despite the fact that it has the strongest propagator. This model has only one viewpoint and this hinders the branching heuristic. Note also that the smallest search trees (but not runtimes) are obtained with the $\forall c$ model that combines the all-different constraint on the primal with the channelling constraints between the primal and dual, when we use both primal and dual variables as decision variables. This combination gives the benefits of the strongest propagator and a dual viewpoint for the branching heuristic.

### 9.2 Quasigroups

| | | QG3(6) | | QG(7) | | QG3(8) | | QG3(9) | |
|---|---|---|---|---|---|---|---|---|---|
| model | heuristic | fails | sec. | fails | sec. | fails | sec. | fails | sec. |
| $\neq$ | SD(p) | 8 | **0.01** | 100 | 0.22 | 1895 | 8.46 | 83630 | 600.61 |
| $\forall$ | SD(p) | 7 | **0.01** | 59 | 0.17 | 955 | 5.76 | **35198** | 385.57 |
| $c$ | SD(p+d) | 7 | 0.02 | 63 | **0.16** | 1117 | 5.81 | 53766 | 463.40 |
| $c$ | SD(p) | 7 | 0.02 | 59 | 0.17 | 1039 | 5.70 | 38196 | 373.38 |
| $c$ | SD(d) | 6 | **0.01** | 54 | 0.19 | 888 | **5.40** | 46539 | 418.96 |
| $c$ | $SD^2$(p+d) | 7 | 0.02 | 63 | 0.17 | 1117 | 5.83 | 53785 | 461.05 |
| $c$ | $SD^2$(p) | 7 | **0.01** | 58 | 0.17 | 1043 | 5.68 | 38198 | 372.41 |
| $c$ | $SD^2$(d) | 6 | **0.01** | 54 | 0.18 | 887 | 5.42 | 46741 | 419.94 |
| $\forall c$ | SD(p+d) | 7 | 0.02 | 54 | **0.16** | 999 | 6.00 | 49678 | 474.82 |
| $\forall c$ | SD(p) | 7 | 0.02 | 59 | 0.18 | 955 | 5.85 | **35198** | 376.06 |
| $\forall c$ | SD(d) | **5** | 0.02 | **52** | 0.2 | 824 | 5.73 | 43278 | 438.81 |
| $\forall c$ | $SD^2$(p+d) | 7 | 0.03 | 54 | 0.17 | 999 | 6.05 | 49702 | 477.04 |
| $\forall c$ | $SD^2$(p) | 7 | 0.02 | 58 | 0.18 | 959 | 5.84 | 35201 | **368.87** |
| $\forall c$ | $SD^2$(d) | **5** | 0.02 | **52** | 0.19 | **823** | 5.80 | 43452 | 432.89 |

Table 6: Number of backtracks (fails) and running time to find all solutions, or prove that there are no solutions, to four instances of QG3 problem. Runtimes are for ILOG Solver 5.300 on 1200MHz, Pentium III processor, and 512 MB of RAM.

The quasigroup existence problem can be modelled as a multiple permutation problem with $2n$ intersecting permutation constraints. We introduce a variable for each entry in the multiplication table of the quasigroup. We then post permutation constraints on the variables of each row and each column. In Tables 6 and 7, we give results for two families of problems. As before, the $\neq$ model gives the worst performance, and by a considerable margin for the larger instances. For QG3, all the other models and branching heuristics give broadly similar performance. A dual viewpoint, either by itself or in combination with the primal viewpoint, does not offer any advantage, but does not hurt much either. For QG4, in Table 7, all the models and branching heuristics are competitive, except for the $\neq$ model and the heuristics that branch only on the dual variables.

### 9.3 Golomb Rulers

To model the Golomb rulers problem as a permutation problem, we introduce a variable for each pairwise distance between marks. Since we may have more values than variables, we introduce additional variables to ensure that there are as many variables as values, as suggested by Geelen (1992). We can then post a permutation constraint on this enlarged set of variables. In Table 8, we give results for finding all optimal length rulers for four instances: Golomb$(n, m)$ means the problem of finding a Golomb ruler of (minimal) length $m$ with $n$ marks. Despite the fact that it has the strongest propagator, the $\forall$ model is not competitive on the larger instances. Model $c$ and heuristic SD(p+d) gives the best runtimes for the larger instances, whereas adding the all-different constraint (model $\forall c$, heuristic SD(p+d)) gives the least search. Being forced to branch on just the primal variables hurts the branching heuristic.

| | | QG4(6) | | QG4(7) | | QG4(8) | | QG4(9) | |
|---|---|---|---|---|---|---|---|---|---|
| model | heuristic | fails | sec. | fails | sec. | fails | sec. | fails | sec. |
| $\neq$ | SD(p) | 6 | **0.01** | 82 | 0.23 | 1779 | 8.29 | 116298 | 843.26 |
| $\forall$ | SD(p) | **4** | **0.01** | **57** | **0.19** | **892** | 5.12 | 52419 | 496.24 |
| $c$ | SD(p+d) | 6 | 0.02 | 59 | 0.20 | 935 | 4.99 | 55232 | 489.89 |
| $c$ | SD(p) | 6 | **0.01** | 59 | 0.20 | 931 | 4.92 | 55397 | 485.72 |
| $c$ | SD(d) | 6 | 0.02 | 74 | 0.21 | 1266 | 7.59 | 83316 | 772.17 |
| $c$ | SD$^2$(p+d) | 6 | 0.02 | 59 | **0.19** | 940 | **4.81** | 55264 | **476.66** |
| $c$ | SD$^2$(p) | 6 | **0.01** | 59 | **0.19** | 936 | 4.87 | 55442 | 478.48 |
| $c$ | SD$^2$(d) | 6 | **0.01** | 73 | 0.22 | 1267 | 7.37 | 82916 | 766.33 |
| $\forall c$ | SD(p+d) | **4** | 0.02 | 57 | **0.19** | 900 | 5.19 | **52045** | 486.72 |
| $\forall c$ | SD(p) | **4** | 0.02 | 57 | 0.20 | **892** | 5.29 | 52419 | 491.54 |
| $\forall c$ | SD(d) | **4** | 0.02 | 67 | 0.21 | 1102 | 7.04 | 73997 | 745.09 |
| $\forall c$ | SD$^2$(p+d) | **4** | **0.01** | 57 | **0.19** | 905 | 5.24 | 52077 | 491.45 |
| $\forall c$ | SD$^2$(p) | **4** | **0.01** | 57 | 0.20 | 897 | 5.23 | 52463 | 493.70 |
| $\forall c$ | SD$^2$(d) | **4** | **0.01** | 66 | 0.23 | 1104 | 7.02 | 73714 | 745.86 |

Table 7: Number of backtracks (fails) and running time to find all solutions, or prove that there are no solutions, to four instances of QG4 problem. Runtimes are for ILOG Solver 5.300 on 1200MHz, Pentium III processor, and 512 MB of RAM.

### 9.4 Sport Scheduling

Unlike the previous problems, we find only the first solution to the sports scheduling problem. This leads to much greater variation in performance between the different models. We report results in Table 9. Good runtimes are obtained with the $c$ and $\forall c$ models, using the dual variables as decision variables, either on their own or in combination with the primal variables.

### 9.5 Magic Squares

We model the order $n$ magic square problem with a $n$ by $n$ matrix of variables which take values from 1 to $n^2$. We then post a permutation constraint on all the variables in the matrix, and sum constraints on the rows, columns and diagonals. Results are given in Table 10. Again, finding just the first solution leads to wide variation in performance between the models. Using only the dual variables as decision variables is a bad choice, but the dual variables are helpful if used as decision variables in combination with the primal variables. For the largest instance solved, the best strategy is the double smallest domain heuristic on model $c$ or model $\forall c$. The former explores a larger search tree, but does so very slightly quicker than the latter.

| model | heuristic | Golomb(7,25) | | Golomb(8,34) | | Golomb(9,44) | | Golomb(10,55) | |
|---|---|---|---|---|---|---|---|---|---|
| | | fails | sec. | fails | sec. | fails | sec. | fails | sec. |
| $\neq$ | SD(p) | 912 | 0.15 | 5543 | 1.12 | – | – | – | – |
| $\forall$ | SD(p) | 500 | **0.11** | 2949 | **0.81** | – | – | – | – |
| $c$ | SD(p+d) | 606 | 0.12 | 3330 | 1.01 | 17002 | **7.54** | 72751 | **49.14** |
| $c$ | SD(p) | 890 | 0.15 | 5343 | 1.25 | – | – | – | – |
| $c$ | SD(d) | 626 | 0.12 | 3390 | 1.02 | 17151 | 7.55 | 73539 | 49.25 |
| $c$ | $SD^2$(p+d) | 608 | 0.12 | 3333 | 1.03 | 17022 | 7.63 | 72853 | 49.37 |
| $c$ | $SD^2$(p) | 928 | 0.17 | 5648 | 1.27 | – | – | – | – |
| $c$ | $SD^2$(d) | 626 | 0.12 | 3390 | 1.03 | 17179 | 7.59 | 73628 | 49.59 |
| $\forall c$ | SD(p+d) | **493** | 0.12 | **2771** | 1.10 | **14313** | 8.29 | **61572** | 54.63 |
| $\forall c$ | SD(p) | 500 | 0.13 | 2949 | 1.08 | – | – | – | – |
| $\forall c$ | SD(d) | 495 | 0.13 | 2782 | 1.10 | 14325 | 8.28 | 61616 | 54.46 |
| $\forall c$ | $SD^2$(p+d) | 504 | 0.14 | 2787 | 1.1 | 14392 | 8.38 | 61898 | 54.94 |
| $\forall c$ | $SD^2$(p) | 542 | 0.14 | 3258 | 1.12 | – | – | – | – |
| $\forall c$ | $SD^2$(d) | 495 | 0.13 | 2794 | 1.11 | 14400 | 8.39 | 61893 | 54.97 |

Table 8: Number of backtracks (fails) and running time to find all optimal solutions to four instances of the Golomb rulers problem, where the optimal length is given. Runtimes are for ILOG Solver 5.300 on 1200MHz, Pentium III processor, and 512 MB of RAM. A dash means that no results were returned after 1 hour.

To conclude, these results show that dynamic branching heuristics can be significantly more effective when they look at both viewpoints of a permutation. Indeed, branching on primal or dual variables was often more important to our results than using a stronger propagator. For example, enforcing GAC on an all-different constraint, and searching just on the primal variables, often gave worse performance than enforcing AC on the channelling constraints, and thus being able to branch on both sets of variables. In addition, in some problem classes, the double smallest domain branching heuristic offered the best performance. As we have argued, this heuristic is consistent with the fail first principle for variable ordering and the succeed first principle for value ordering.

It is worth noting that the results of our experiments run counter to the usual expectations of value ordering. We found that double smallest domain (that is, smallest domain for both variable ordering and value ordering) gave different numbers of backtracks to smallest domain variable ordering, even when finding all solutions. It is generally thought that value ordering makes no difference to the overall search effort when finding all solutions, if chronological backtracking is used. Indeed, the argument given earlier for succeed first as a value ordering principle is based on finding only one solution: if we choose the right value, we can avoid backtracking to choose another one. If we want to find all solutions, we shall have to backtrack to try all the alternative values anyway. Smith (2000) shows how value ordering can make a difference to the search in Langford's problem, even when finding all solutions. In brief, when we backtrack having tried the assignment $Var = value$, we can post the constraint $Var \neq value$. In some cases, propagation may now lead to immediate failure. A good ordering for the values can therefore save search.

| | | Sport(6) | | Sport(8) | | Sport(10) | | Sport(12) | |
|---|---|---|---|---|---|---|---|---|---|
| model | heuristic | fails | sec. | fails | sec. | fails | sec. | fails | sec. |
| $\neq$ | SD(p) | **0** | **0.00** | 1248 | 0.22 | 1863275 | 397.70 | 5777382 | 1971.92 |
| $\forall$ | SD(p) | **0** | 0.01 | 566 | 0.15 | 1361686 | 350.92 | 3522705 | 1444.44 |
| $c$ | SD(p+d) | 624 | 0.09 | 4 | **0.01** | **7** | **0.03** | 5232 | **1.78** |
| $c$ | SD(p) | **0** | **0.00** | 566 | 0.14 | 1376143 | 355.99 | 3537447 | 1368.84 |
| $c$ | SD(d) | 589 | 0.07 | **3** | **0.01** | 336 | 0.07 | 6368 | 1.9 |
| $c$ | SD$^2$(p+d) | 7 | **0.00** | 9 | **0.01** | 1112 | 0.30 | 46122 | 18.4 |
| $c$ | SD$^2$(p) | 113 | 0.02 | 6601 | 0.94 | 820693 | 168.91 | – | – |
| $c$ | SD$^2$(d) | 514 | 0.06 | 43 | **0.01** | 7028 | 1.58 | 6252 | 2.29 |
| $\forall c$ | SD(p+d) | 624 | 0.10 | 4 | **0.01** | **7** | **0.03** | **5190** | 1.98 |
| $\forall c$ | SD(p) | **0** | 0.01 | 566 | 0.16 | 1361686 | 372.10 | 3522705 | 1495.41 |
| $\forall c$ | SD(d) | 589 | 0.09 | **3** | **0.01** | 329 | 0.08 | 6262 | 2.18 |
| $\forall c$ | SD$^2$(p+d) | 7 | **0.00** | 9 | **0.01** | 1102 | 0.35 | 45125 | 20.98 |
| $\forall c$ | SD$^2$(p) | 113 | 0.02 | 6563 | 1.09 | 812696 | 186.23 | – | – |
| $\forall c$ | SD$^2$(d) | 514 | 0.07 | 43 | 0.02 | 6920 | 1.76 | 6129 | 2.55 |

Table 9: Number of backtracks (fails) and running time to find the first solution to four instances of the sports scheduling problem. Runtimes are for ILOG Solver 5.300 on 1200MHz, Pentium III processor, and 512 MB of RAM.

## 10. Injective Mappings

In many problems, variables may be constrained to take unique values, but we have more values than variables. That is, we are looking for an injective mapping from the variables to the values. For example, an optimal 5-tick Golomb ruler has ticks at the marks 0, 1, 4, 9, and 11. The 10 inter-tick distances are all different but do not form a permutation as the distance 6 is absent. Finding a 5-tick Golomb ruler of length 11 can be modelled as a permutation problem by introducing an additional 11th variable to take on the missing value 6. Indeed, this is the method we use to model the problem in the last section. However, there are a number of alternative ways to model an injection from $n$ variables into $m$ values which we explore here.

For example, there are two simple primal models of an injection. In each we have $n$ primal variables which take one of $m$ possible values. In the primal all-different model (denoted by "$\forall$"), we simple post a single all-different constraint on the primal variables.

| | | Magic(3) | | Magic(4) | | Magic(5) | | Magic(6) | |
|---|---|---|---|---|---|---|---|---|---|
| model | heuristic | fails | sec. | fails | sec. | fails | sec. | fails | sec. |
| $\neq$ | SD(p) | 6 | **0.00** | 20 | **0.00** | 1576 | 0.11 | – | – |
| $\forall$ | SD(p) | **4** | **0.00** | 19 | **0.00** | 1355 | 0.11 | 2748609 | 196.45 |
| $c$ | SD(p+d) | 5 | **0.00** | 18 | **0.00** | 4637 | 0.37 | – | – |
| $c$ | SD(p) | **4** | **0.00** | 20 | **0.00** | 1457 | 0.14 | 3448162 | 249.84 |
| $c$ | SD(d) | 5 | **0.00** | 37 | 0.01 | 49312 | 4.61 | – | – |
| $c$ | $SD^2$(p+d) | 5 | **0.00** | 10 | **0.00** | 555 | 0.06 | 463865 | **37.41** |
| $c$ | $SD^2$(p) | **4** | **0.00** | 11 | **0.00** | 495 | **0.05** | 1648408 | 132.35 |
| $c$ | $SD^2$(d) | 5 | **0.00** | 18 | **0.00** | 928217 | 86.07 | – | – |
| $\forall c$ | SD(p+d) | 5 | 0.01 | 18 | **0.00** | 4436 | 0.48 | – | – |
| $\forall c$ | SD(p) | **4** | **0.00** | 19 | **0.00** | 1355 | 0.17 | – | – |
| $\forall c$ | SD(d) | 5 | **0.00** | **5** | **0.00** | 42426 | 5.33 | – | – |
| $\forall c$ | $SD^2$(p+d) | 5 | 0.02 | 10 | 0.01 | 435 | 0.07 | **290103** | 39.01 |
| $\forall c$ | $SD^2$(p) | **4** | **0.00** | 11 | **0.00** | **355** | **0.05** | 1083993 | 148.73 |
| $\forall c$ | $SD^2$(d) | 5 | **0.00** | 16 | **0.00** | 919057 | 106.55 | – | – |

Table 10: Number of backtracks (fails) and running time to find the first solution to four instances of magic square problem. Runtimes are for ILOG Solver 5.300 on 1200MHz, Pentium III processor, and 512 MB of RAM. A dash means that no results were returned after 1 hour.

In the primal not-equals model (denoted by "$\neq$") we post binary not-equals constraints between every two distinct primal variables. We can also use dual models. For example, in the dual not-equals model, we have $m$ dual variables, each with a domain of $m$ possible values ($m - n$ of these are dummy values), and binary not-equals constraints between each pair of dual variables.

We will consider three different combined models which channel between primal and dual models. In the first combined model (denoted by "$c_1$"), we have channelling constraints of the form $x_i = j$ implies $d_j = i$ and no additional dummy values for the dual variables. In the second combined model (denoted by "$c_2$"), the dual variables have $m - n$ extra dummy values, and we have channelling constraints of the form $x_i = j$ iff $d_j = i$. In the third combined model (denoted by "$c_3$"), the dual variables have just a single extra dummy value, and we have channelling constraints of the form $x_i = j$ iff $d_j = i$ but only when $j$ is not equal to the dummy value. Note that any of these channelling constraints alone (without additional constraints on the primal or dual variables) is enough to define an injection.

We can also model an injection by introducing $m - n$ dummy primal variables and ensuring that this extended set of variables forms a bijection. This case is, however, covered by our earlier results on permutations.

### 10.1 Arc-Consistency

We first prove that, with respect to arc-consistency, the first type of channelling constraints are as tight as the primal not-equals constraints, but less tight than the primal all-different constraint. Then, we prove that the second type of channelling constraints are as tight as the primal not-equals constraints, but less tight than the channelling and dual not-equals constraints, which are less tight than the primal all-different constraint. Finally, we prove that the third type of channelling constraints are as tight as the primal not-equals constraints but less tight than the primal all-different constraint. This means that the three types of channelling constraints give the same pruning when we enforce arc-consistency as the primal not-equals constraints. Note, however, that we get more pruning when we add the dual not-equals constraints (but not the primal not-equals constraints). This is different to permutations where neither the addition of the primal nor the dual not-equals constraints to the channelling constraint gave more pruning.

**Theorem 11** *On an injection problem:*

$$GAC_{\forall} \rightarrow AC_{\neq c_1} \leftrightarrow AC_{c_1} \leftrightarrow AC_{\neq}$$

**Proof:** To show $\text{GAC}_{\forall} \rightarrow \text{AC}_{c_1}$, consider an injection problem whose primal all-different constraint is GAC. Suppose the channelling constraint between $x_i$ and $d_j$ was not AC. Then $x_i$ is set to $j$ and $d_j$ has $i$ eliminated from its domain. But this is not possible by the construction of the primal and dual model. Hence the channelling constraints are all AC. To show strictness, consider an injection problem in which $x_1 = x_2 = x_3 = \{1, 2\}$ and $d_1 = d_2 = d_3 = d_4 = \{1, 2, 3\}$. This is $\text{AC}_{c_1}$ but not $\text{GAC}_{\forall}$.

To show $\text{AC}_{c_1} \leftrightarrow \text{AC}_{\neq}$, suppose that the channelling constraints are AC. Consider a not-equals constraint, $x_i \neq x_j$ (where $i \neq j$) that is not AC. Now, $x_i$ and $x_j$ must have the same singleton domain, $\{k\}$. Consider the channelling constraint between $x_i$ and $d_k$. The only AC value for $d_k$ is $i$. Similarly, the only AC value for $d_k$ in the channelling constraint between $x_j$ and $d_k$ is $j$. But $i \neq j$. Hence, $d_k$ has no AC values. This is a contradiction as the channelling constraints are AC. Hence all not-equals constraints are AC. Now suppose that the not-equals constraints are AC. Consider a channelling constraint between $x_i$ and $d_j$ that is not AC. Then $x_i$ is set to $j$ and $d_j$ has $i$ eliminated from its domain. But for $i$ to be eliminated from the domain of $d_j$, some other primal variable, say $x_k$ where $k \neq i$, is set to $j$, which eliminate $j$ from the domain of $x_i$ (since the not-equals constraints are AC). Hence, it is not possible to set $x_i$ to $j$ and $d_j$ has $i$ eliminated from its domain. Thus, all channelling constraints are AC. □

**Theorem 12** *On an injection problem:*

$$GAC_{\forall} \rightarrow AC_{\neq c_2 \neq} \leftrightarrow AC_{c_2 \neq} \rightarrow AC_{c_2} \leftrightarrow AC_{\neq}$$

**Proof:** To show $\text{GAC}_{\forall} \rightarrow \text{AC}_{c_2 \neq}$, consider an injection problem which is $\text{GAC}_{\forall}$. Suppose the not-equal constraint between $d_i$ and $d_j$ was not AC. Then, in the first case, $d_i = d_j = k$ and $k < n+1$, which is impossible because the channelling constraints $x_k = i$ iff $d_i = k$ and $x_k = j$ iff $d_j = k$ are AC. In the second case, $k$ would be greater than $n$, which is impossible by construction of the primal and dual model. Hence all binary not-equal constraints on the dual variables are AC. To show strictness, consider an injection in which $x_1 = x_2 = x_3 = \{1,2\}$, $d_1 = d_2 = \{1,2,3,4,5\}$, and $d_3 = d_4 = d_5 = \{4,5\}$. This is $\text{AC}_{c2\neq_d}$ but not $\text{GAC}_\forall$.

To show $\text{AC}_{c2\neq} \rightarrow \text{AC}_{c2}$, by monotonicity, we have $\text{AC}_{c2\neq} \hookrightarrow \text{AC}_{c2}$. To show strictness, consider an injection problem in which $x_1 = x_2 = x_3 = \{1,2\}$, and $d_1 = d_2 = \{1,2,3,4\}$, and $d_3 = d_4 = \{4\}$. This is $\text{AC}_{c_2}$ but not $\text{GAC}_{c2\neq}$.

To show $\text{AC}_{c_2} \leftrightarrow \text{AC}_{\neq}$, suppose that the channelling constraints are AC. Consider a not-equals constraint, $x_i \neq x_j$ (where $i \neq j$) that is not AC. Now, $x_i$ and $x_j$ must have the same singleton domain, $\{k\}$. Consider the channelling constraint between $x_i$ and $d_k$. The only AC value for $d_k$ is $i$. Similarly, the only AC value for $d_k$ in the channelling constraint between $x_j$ and $d_k$ is $j$. But $i \neq j$. Hence $d_k$ has no AC values. This is a contradiction as the channelling constraints are AC. Hence all not-equals constraints are AC. To show the reverse, suppose that the not-equals constraints are AC. Consider a channelling constraint, $x_i = j$ iff $d_j = i$, that is not AC. Then, either $x_i$ is set to $j$ and $d_j$ has $i$ eliminated from its domain, or $d_j$ is set to $i$ and $x_i$ has $j$ eliminated from its domain. But, for $i$ to be eliminated from the domain of $d_j$, some other primal variable, say $x_k$ where $k \neq i$, is set to $j$, which will eliminate $j$ from the domain of $x_i$ (since the not-equals constraints are AC). Hence it is not possible to set $x_i$ to $j$ and $d_j$ has $i$ eliminated from its domain. For $d_j$ to be set to $i$, all the other values must be removed from its domain, but there is no way to remove any of the values bigger than $n$ from the domain of $d_j$, because at most we have $n$ primal variables. Thus, all channelling constraints are AC. □

**Theorem 13** *On an injection problem:*

$$GAC_\forall \rightarrow AC_{c_3} \leftrightarrow AC_{\neq}$$

**Proof:** To show $\text{GAC}_\forall \rightarrow \text{AC}_{c_3}$, consider an injection in which $x_1 = x_2 = x_3 = \{1,2\}$, $x_4 = \{1,2,3,4,5\}$, $d_1 = d_2 = \{1,2,3,4,5\}$, and $d_3 = d_4 = d_5 = \{4,5\}$. This is $\text{GAC}_{c3|W|}$, but not $\text{GAC}_\forall$.

To show $\text{AC}_{c_3} \leftrightarrow \text{AC}_{\neq}$, suppose that the channelling constraints are AC. Consider a not-equals constraint, $x_i \neq x_j$ (where $i \neq j$) that is not AC. Now, $x_i$ and $x_j$ must have the same singleton domain, $\{k\}$. Consider the channelling constraint between $x_i$ and $d_k$. The only AC value for $d_k$ is $i$. Similarly, the only AC value for $d_k$ in the channelling constraint between $x_j$ and $d_k$ is $j$. But $i \neq j$. Hence $d_k$ has no AC values. This is a contradiction as the channelling constraints are AC. Hence all not-equals constraints are AC. To show the reverse, suppose that the not-equals constraints are AC. Consider a channelling constraint, $x_i = j$ iff $d_j = i$, that is not AC. Then, either $x_i$ is set to $j$ and $d_j$ has $i$ eliminated from its domain, or $d_j$ is set to $i$ and $x_i$ has $j$ eliminated from its domain. But, for $i$ to be eliminated from the domain of $d_j$, some other primal variable, say $x_k$ where $k \neq i$, is set to $j$, which will eliminate $j$ from the domain of $x_i$ (since the not-equals constraints are AC). Hence it is not possible to set $x_i$ to $j$ and $d_j$ has $i$ eliminated from its domain. For $d_j$ to be set to $i$, all the other values must be removed from its domain, but there is no way to remove any of the values bigger than $n$ from the domain of $d_j$, because we have at most $n$ primal variables. Thus, all channelling constraints are AC. □

### 10.2 Asymptotic Comparison

The previous results compare the different models with respect to the amount of pruning achieved. We can, for example, now rule out a model like "$\neq c_1$" when enforcing AC since we get just as much pruning at less cost on the model $c_1$. However, these results do not distinguish between, say, a model with primal not-equals constraints, or any of the combined models $c_1$, $c_2$ or $c_3$. We get the same pruning in all four. We can add some details to these results by comparing the asymptotic behaviour.

The relative cost of achieving $\text{GAC}_\forall$ is $O(n^2m^2)$, where $n$ is the number of variables and $m$ is their domain size. $\text{AC}_{c_1}$, $\text{AC}_{c_2}$, and $\text{AC}_{c_3}$ naively take $O(nm^3)$ time. However, by taking advantage of the functional nature of channelling constraints, we can reduce this to $O(nm^2)$ for $c_2$ and $c_3$ and $O(nm)$ for $c_1$. We proved in Theorem 11 that $\text{GAC}_\forall \rightarrow \text{AC}_{c_1} \leftrightarrow \text{AC}_{\neq}$ and their costs are $O(n^2m^2)$, $O(nm)$, and $O(n^2)$ respectively. Asymptotic analysis shows that enforcing $\text{AC}_{c_1}$ has asymptotically slightly more cost than enforcing $\text{AC}_{\neq}$. However, having the dual variables could be advantageous in conjunction with variable and value ordering heuristics. We also proved in Theorem 12 that $\text{GAC}_\forall \rightarrow \text{AC}_{c_2\neq} \rightarrow \text{AC}_{c_2} \leftrightarrow \text{AC}_{\neq}$ and their costs are $O(n^2m^2)$, $O(nm^2)$, $O(nm^2)$, and $O(n^2)$ respectively. Asymptotic analysis shows that the channelling constraints are more costly than the not-equals constraints and bring no more pruning. When we add not-equals constraints on the dual variables, the overall asymptotic cost is still the same as the channelling constraints alone, but we achieve more pruning. It is therefore a model worth considering. Finally, in Theorem 13 we proved that $\text{GAC}_\forall \rightarrow \text{AC}_{c_3} \leftrightarrow \text{AC}_{\neq}$ and their costs are $O(n^2m^2)$, $O(nm^2)$, and $O(n^2)$ respectively. Again, asymptotic analysis shows that channelling constraints are more costly than the not-equals constraints and bring no more pruning. Maintaining generalised arc-consistency on the all-different constraint is again the most costly.

To conclude, these results show that, as might be expected, we in general get more pruning if we increase the asymptotic cost. Models worth considering are the primal not-equals model, $c_2 \neq$, and the primal all-different model. Each gives a different amount of pruning at a different asymptotic cost. We might also consider $c_1$ instead of the primal not-equals model since, whilst it is asymptotically slightly more expensive, it lets us branch on dual variables.

### 10.3 Experiments With Static Orderings

We again ran some experiments to explore the significance of these theoretical and asymptotic differences. Table 11 gives results on some instances of the Golomb rulers problem using a static variable ordering. The experiments are again consistent with the theoretical results. First, enforcing GAC on an all-different constraint achieves the most pruning and has the smallest runtimes. Second, on these problems instances, enforcing AC on the binary not-equals constraints achieves the same amount of pruning as maintaining AC on the channelling constraints. In addition, enforcing AC on the channelling constraints takes longer to achieve. Third, adding the channelling constraints to the primal all-different constraint does not increase pruning, and merely adds overhead to the runtime.

| | | Golomb(8,34) | | Golomb(9,44) | | Golomb(10,55) | | Golomb(11,72) | |
|---|---|---|---|---|---|---|---|---|---|
| model | heuristic | fails | sec. | fails | sec. | fails | sec. | fails | sec. |
| $\forall$ | static | **82** | **0.02** | **724** | **0.26** | **3461** | **2.08** | **18493** | **13.63** |
| $c_2$ | static | 104 | 0.03 | 1110 | 0.38 | 7122 | 3.46 | 37404 | 23.02 |
| $\neq$ | static | 104 | 0.03 | 1110 | 0.34 | 7122 | 3.03 | 37404 | 20.32 |
| $\forall c_2$ | static | **82** | 0.03 | **724** | 0.36 | **3461** | 2.76 | **18493** | 17.97 |

Table 11: Number of backtracks (fails) and running time to find the first solution to four instances of the Golomb rulers problem. Runtimes are for ILOG Solver 5.300 on 1200MHz, Pentium III processor, and 512 MB of RAM.

### 10.4 Dynamic Variable And Value Ordering Heuristics

| | | Golomb(8,34) | | Golomb(9,44) | | Golomb(10,55) | | Golomb(11,72) | |
|---|---|---|---|---|---|---|---|---|---|
| model | heuristic | fails | sec. | fails | sec. | fails | sec. | fails | sec. |
| $\neq$ | SD(p) | 326 | 0.06 | 3810 | 0.96 | 50526 | 16.67 | 800169 | 352.8 |
| $\forall$ | SD(p) | 238 | 0.04 | 2629 | 0.75 | 32705 | 13.12 | 563011 | 266.52 |
| $c_2$ | SD(p+d) | 11 | **0.00** | 2010 | 0.57 | 2288 | 0.86 | 982 | 0.48 |
| $c_2$ | SD(p) | 326 | 0.07 | 3810 | 1.13 | 50526 | 20.42 | 800169 | 418.03 |
| $c_2$ | SD(d) | 12 | **0.00** | 2333 | 0.61 | 2822 | 0.90 | 1254 | 0.52 |
| $c_2$ | $SD^2$(p+d) | 12 | 0.01 | 2033 | 0.58 | 2374 | 0.86 | 984 | 0.48 |
| $c_2$ | $SD^2$(p) | 335 | 0.06 | 4244 | 1.18 | 57158 | 21.54 | 898457 | 441.15 |
| $c_2$ | $SD^2$(d) | 12 | **0.00** | 2342 | 0.60 | 2911 | 0.91 | 1247 | 0.51 |
| $\forall c_2$ | SD(p+d) | **10** | **0.00** | **904** | 0.44 | **1076** | 0.66 | 598 | **0.43** |
| $\forall c_2$ | SD(p) | 238 | 0.07 | 2629 | 1.10 | 32705 | 19.32 | 563011 | 419.45 |
| $\forall c_2$ | SD(d) | 11 | **0.00** | 906 | 0.44 | 1087 | **0.64** | 605 | 0.44 |
| $\forall c_2$ | $SD^2$(p+d) | **10** | **0.00** | 914 | **0.43** | 1125 | 0.69 | **588** | 0.44 |
| $\forall c_2$ | $SD^2$(p) | 254 | 0.07 | 3054 | 1.17 | 39143 | 21.21 | 663896 | 456.75 |
| $\forall c_2$ | $SD^2$(d) | 11 | 0.01 | 909 | **0.43** | 1131 | 0.70 | 592 | 0.44 |

Table 12: Number of backtracks (fails) and running time to find the first solution to four instances of the Golomb rulers problem. Runtimes are for ILOG Solver 5.300 on 1200MHz, Pentium III processor, and 512 MB of RAM.

We also explored the advantage of multiple viewpoints of injection problems for dynamic variable and value ordering heuristics. In Table 12, we give results for Golomb ruler problems. We observe that the primal all-different model is not competitive on the larger problems. The best runtimes are obtained with the channelling constraints (and a primal all-different constraint) using the smallest domain or the double smallest domain heuristic on both sets of variables or on the dual variables. Being forced to branch on just the primal variables hurts the branching heuristic. A dual viewpoint appears to offer the branching heuristic very significant advantages on this problem.

| | | Sport(7) | | Sport(9) | | Sport(11) | |
|---|---|---|---|---|---|---|---|
| model | heuristic | fails | sec. | fails | sec. | fails | sec. |
| $\neq$ | SD(p) | 14 | **0.00** | 140287 | 15.33 | – | – |
| $\forall$ | SD(p) | 14 | **0.00** | 138643 | 16.12 | – | – |
| $c_2$ | SD(p+d) | 3 | **0.00** | 34 | **0.01** | 43877 | **8.04** |
| $c_2$ | SD(p) | 14 | **0.00** | 140294 | 17.21 | – | – |
| $c_2$ | SD(d) | **0** | **0.00** | 33 | **0.01** | 1829954 | 268.73 |
| $c_2$ | $SD^2$(p+d) | 3 | **0.00** | 4535 | 0.67 | 910362 | 185.63 |
| $c_2$ | $SD^2$(p) | 14 | **0.00** | 143989 | 17.71 | – | – |
| $c_2$ | $SD^2$(d) | 2 | **0.00** | 11424 | 1.36 | 12536523 | 1787.21 |
| $\forall c_2$ | SD(p+d) | 3 | **0.00** | 28 | **0.01** | **38555** | 9.05 |
| $\forall c_2$ | SD(p) | 14 | 0.01 | 138643 | 20.27 | – | – |
| $\forall c_2$ | SD(d) | **0** | **0.00** | **31** | 0.02 | 374829 | 78.53 |
| $\forall c_2$ | $SD^2$(p+d) | 3 | **0.00** | 2013 | 0.34 | 600686 | 151.19 |
| $\forall c_2$ | $SD^2$(p) | 14 | **0.00** | 142313 | 20.31 | – | – |
| $\forall c_2$ | $SD^2$(d) | 2 | **0.00** | 3238 | 0.52 | 1854082 | 431.19 |

Table 13: Number of backtracks (fails) and running time to find the first solution to three instances of sport scheduling problem. Runtimes are for ILOG Solver 5.300 on 1200MHz, Pentium III processor, and 512 MB of RAM. A dash means no solution is found after 1 hour.

In Table 13, we give results for the sport scheduling problem when there are an odd number of weeks. Despite the fact that it has the strongest propagator, the primal all-different model is not competitive on the larger problems. The best runtimes are obtained with the channelling constraints and branching on the primal or dual variable with smallest domain. As with the Golomb ruler problem, being forced to branch on just the primal variables hurts the branching heuristic. A dual viewpoint appears to offer the branching heuristic very significant advantages on this problem. Note also that on the largest instance, the smallest search tree is obtained with the channelling and the all-different constraints, branching on the primal or dual variable with smallest domain. To conclude, dynamic branching heuristics can again be significantly more effective when they look at both the primal and dual viewpoint.

## 11. Related Work

Cheng et al. (1999) studied modelling and solving the $n$-queens problem, and a nurse rostering problem using channelling constraints. They show that channelling constraints increase the amount of constraint propagation. They conjecture that the overheads associated with channelling constraints will pay off on problems which require large amounts of search, or lead to thrashing behaviour. They also show that channelling constraints open the door to interesting value ordering heuristics. For permutation problems, a similar idea was previously proposed by Geelen (1992).

Choi and Lee (2002) focused on the study of combined models of permutation problems. Their study included not only the permutation constraints, but also all the other constraints of the problem. Their comparison measure is an extension of the propagator comparison approach of Schulte and Stuckey (2001), which measures the different combined models with respect to their ability to prune the search space with constraint propagation. However, their measure is independent of the level of consistency maintained on the constraints and depends upon the set of correct propagators instead. They theoretically discover the criteria under which minimal combined models have the same pruning power as full combined models and empirically demonstrate the results on different permutation problems.

Bacchus et al. (2002) formally studied the effectiveness of two modelling techniques that transform a non-binary CSP into an equivalent binary CSP, namely, the dual transformation and the hidden one. An original model of the problem, its dual and its hidden transformations are compared with respect to the performance of a number of local consistency techniques including arc-consistency, and with respect to the chronological backtracking algorithm, FC, and MAC.

Borret and Tsang (1999) developed a framework for systematic model selection. They demonstrated their approach on the evaluation of adding a certain class of implied constraints to an original model. The evaluation heuristic used is based on an extension of the theoretical complexity estimates proposed by Nadel (1990). Their experimental results show that the approach is promising. However, with this approach one needs the instance data to be an explicit input to the methods.

## 12. Conclusions

We have performed an extensive study of dual modelling on permutation and injection problems. To compare models, we defined a measure of constraint tightness parameterized by the level of local consistency being enforced. For permutation problems and enforcing arc-consistency, we proved that a single primal all-different constraint is tighter than channelling constraints, but that channelling constraints are tighter than primal not-equals constraints. The reason for this difference is that the primal not-equals constraints detect singleton variables (i.e. those variables with a single value), the channelling constraints detect singleton variables and singleton values (i.e. those values which occur in the domain of a single variable), whilst the primal all-different constraint detects global consistency (which includes singleton variables, singleton values and many other situations). For lower levels of local consistency (e.g. that maintained by forward checking), channelling constraints remain tighter than primal not-equals constraints. However, for certain higher levels of local consistency like path inverse consistency, channelling constraints are incomparable to primal not-equals constraints. For injection problems, we proved that, with respect to arc-consistency, a single primal all-different constraint is tighter than channelling constraints together with the dual not-equals constraints, but that the channelling constraints alone are as tight as the primal not-equals constraints. The asymptotic analysis allowed us to reduce further the number of models that might be worth considering. Experimental results on a wide range of problems supported these theoretical results. For example, adding binary not-equals constraints to the channelling constraints does not increasing pruning, and merely adds overhead to the runtimes. However, the experimental results also demonstrated the very significant benefits of being able to branch on both primal and dual variables. In many cases, we obtained the best runtimes with just channelling constraints and a branching heuristic that looked at both primal and dual viewpoints.

What general lessons can be learnt from this study? First, there are many possible models of even a simple problem like finding a permutation or an injection. In addition, no one model is best in all situations. We therefore need to support the user in modelling even simple problems. Second, it often pays to construct redundant models with multiple viewpoints of the same problem. Despite the overheads, the ability to branch on dual variables can be very beneficial. Branching heuristics that consider multiple viewpoints can be very effective. Third, the additional constraint propagation provided by global constraints like all-different may not justify their cost. We often saw better performance when we threw out the all-different constraint. Fourth, our measure of constraint tightness can be used to compare different constraint models. However, this measure can only reject certain models on the basis that they add overhead. We still must run experiments to determine if the additional constraint propagation provided by tighter models is worth the cost of this constraint propagation. Ultimately, the question being addressed is central to many problems in artificial intelligence: the trade-off between search and inference.

## Acknowledgements

B. Hnich and T. Walsh are currently supported by Science Foundation Ireland (SFI) and an ILOG software grant. T. Walsh was also supported by an EPSRC advanced research fellowship. We thank the other members of the APES research group (http://www.dcs.st-and.ac.uk/~apes) for helpful discussions, and especially Ian Gent who encouraged us to write this paper.